\documentclass[10pt,twocolumn,letterpaper]{article}

\usepackage[pagenumbers]{wacv} 
\usepackage{adjustbox}
\usepackage{multirow}
\usepackage{siunitx}
\DeclareSIUnit\pixel{px}
\usepackage{pgfplots}
\pgfplotsset{compat=1.18}   
\usepgfplotslibrary{groupplots}
\usetikzlibrary{plotmarks}  
\usepackage{amsmath}
\usepackage{amssymb}
\usepackage{mathtools} 
\usepackage{xcolor}
\usepackage[table]{xcolor}   
\usepackage{colortbl} 
\newcommand{\bmfirst}[1]{\cellcolor{blue!25}\textbf{#1}}
\newcommand{\bmsecond}[1]{\cellcolor{blue!18}{#1}}
\newcommand{\bmthird}[1]{\cellcolor{blue!8}{#1}}

\newcommand{\rdfirst}[1]{\cellcolor{red!25}\textbf{#1}}
\newcommand{\rdsecond}[1]{\cellcolor{red!18}{#1}}
\newcommand{\rdthird}[1]{\cellcolor{red!8}{#1}}

\usepackage{tikz}
\usetikzlibrary{positioning,arrows.meta,fit,calc,shapes.misc}
\def\checkmark{\tikz\fill[scale=0.4](0,.35) -- (.25,0) -- (1,.7) -- (.25,.15) -- cycle;}

\definecolor{wacvblue}{rgb}{0.21,0.49,0.74}
\usepackage[pagebackref,breaklinks,colorlinks,allcolors=wacvblue]{hyperref}
\usepackage{amsmath}

\def\wacvPaperID{3254} 
\def\confName{WACV}
\def\confYear{2026}

\title{Learning Subglacial Bed Topography from Sparse Radar with Physics-Guided Residuals}

\author{Bayu Adhi Tama$^{1}$ \and Jianwu Wang$^{2}$ \and Vandana Janeja$^{2}$ \and Mostafa Cham$^{2}$\\
$^{1}$iHARP, University of Maryland Baltimore County (UMBC)\\
$^{2}$Department of Information Systems, University of Maryland Baltimore County (UMBC)\\
{\tt\small \{bayu; jianwu, vjaneja, mcham2\}@umbc.edu}
}

\begin{document}
\maketitle

\begin{abstract}
Accurate subglacial bed topography is essential for ice-sheet modeling, yet radar observations are sparse and uneven. We propose a physics-guided residual learning framework that predicts bed \emph{thickness residuals} over a BedMachine prior and reconstructs bed from the observed surface. A DeepLabV3$+$ decoder over a standard encoder (e.g., ResNet-50) is trained with lightweight physics and data terms: multi-scale mass conservation, flow-aligned total variation, Laplacian damping, non-negativity of thickness, a ramped prior-consistency term, and a masked Huber fit to radar picks modulated by a confidence map. To measure real-world generalization, we adopt leakage-safe \emph{block-wise} hold-outs (vertical/horizontal) with safety buffers and report metrics only on held-out cores. Across two Greenland sub-regions, our approach achieves strong test-core accuracy (RMSE $3.05$–$10.54$\,m; $R^2=0.993$–$0.999$) and high structural fidelity (SSIM $\ge 0.998$, PSNR up to $52.9$\,dB), outperforming U-Net, Attention U-Net, FPN, and a plain CNN. The residual-over-prior design, combined with physics, yields spatially coherent, physically plausible beds suitable for operational mapping under domain shift.\vspace{-4pt}
\end{abstract}

\section{Introduction}
\label{sec:intro}



Understanding the bed beneath ice sheets is central to modeling ice flow, basal hydrology, and sea-level rise~\cite{ipcc2023}. Yet airborne radar sounding is sparse and heterogeneous—dense along flight lines and absent across large interiors—so naive interpolation is unreliable and often fails to generalize across regions~\cite{barnes2021transferability,aakesson2021future}. Community products such as BedMachine~\cite{morlighem2017bedmachine,morlighem2014deeply} provide a valuable prior assembled from diverse data and assumptions, but they may be locally biased or overly smooth where constraints are weak. A practical approach should (i) exploit the prior to capture broad structure, (ii) learn data-driven corrections that respect physics, and (iii) be evaluated under protocols that reflect deployment to truly unseen spatial extents~\cite{bamber2013new,bamber2001new}.

We address this with a \emph{physics-guided residual learning} framework. Instead of regressing absolute bed elevation, we predict a normalized \emph{residual thickness} with respect to a BedMachine-derived prior and recover bed via the observed surface. This residual-over-prior design delegates large-scale structure to the prior while focusing learning capacity on physically plausible corrections. We instantiate the predictor with a DeepLabV3$+$~\cite{chen2018encoder} decoder over standard encoders (e.g., ResNet-50~\cite{he2016deep}) and regularize training using lightweight, scalable physics and data terms: (1) a multi-scale mass-conservation penalty~\cite{morlighem2011mass} on thickness tendency and advective flux, (2) a flow-aligned total variation that down-weights along-flow gradients relative to cross-flow, (3) a Laplacian/high-pass penalty to suppress oscillations, (4) a non-negativity hinge on thickness, (5) a ramped prior-consistency term emphasized where radar confidence is low, and (6) a radar data term that fits residual thickness at pick locations via a masked Huber loss. Stabilizers—Exponential Moving Average (EMA) weights, test-time augmentation (TTA), and seam-free tiled inference—improve robustness for full-grid predictions.

Equally important is how we measure generalization. Random patch splits can leak context across train/test via large receptive fields and yield optimistic estimates. We therefore adopt two orthogonal \emph{block-wise} hold-outs—vertical (west train / east test) and horizontal (south train / north test)—together with a safety buffer that carves out a held-out \emph{test core} fully isolated from training tiles. All metrics are reported only on this test core. Across two Greenland sub-regions (e.g., Upernavik Isstr{\o}m) and both hold-outs, our physics-guided residual model consistently outperforms U-Net, Feature Pyramid Network (FPN), and plain CNN baselines, producing spatially coherent, physically plausible beds suitable for operational mapping under domain shift and interior gaps. In summary, our contributions are:
\begin{itemize}
\item A residual-over-prior formulation coupled with mass-conservation and flow-aware regularization for subglacial bed mapping.
\item A leakage-safe evaluation protocol: orthogonal block-wise hold-outs (west/east, south/north) with receptive-field buffers; metrics reported on held-out test cores.
\item State-of-the-art performance against U-Net, FPN, and CNN baselines across both block-wise splits.
\end{itemize}

\begin{table}[ht!]
\caption{\textbf{Feature comparison with recent methods.} We contrast our residual-over-prior design against DeepTopoNet~\cite{tama2025deeptoponet} and GraphTopoNet~\cite{tama2025improving}.}.
    \label{tab:sota}
    \centering
    \begin{adjustbox}{width=0.48\textwidth}
    \begin{tabular}{p{10em}ccc}
    \hline    Feature&Ours&DeepTopoNet~\cite{tama2025deeptoponet}&GraphTopoNet~\cite{tama2025improving}\\
    \hline
    $\mathcal{L}_{prior}$&Thickness&Bed&Bed\\
    Radar confidence&\checkmark&$\times$&\checkmark\\
    Physics-guided&\checkmark&$\times$&$\times$\\    
    Non-negativity of thickness&\checkmark&$\times$&$\times$\\
    Spatial-aware validation&\checkmark&$\times$&$\times$\\
    Leakage-safe evaluation&\checkmark&$\times$&$\times$\\ 
    \hline
    \end{tabular} 
    \end{adjustbox}
\end{table}

\section{Related Work}
\noindent\textbf{Spatial interpolation.}
Classical spatial interpolation~\cite{li2000comparison,lam1983spatial} infers values at unsampled locations from nearby observations. Deterministic schemes such as inverse distance weighting (IDW)~\cite{li2008review} and triangulated irregular networks (TIN)~\cite{feng2024critical} assume simple, pre-defined spatial relationships. Geostatistical approaches—most notably kriging and its variants~\cite{wackernagel2003ordinary,wackernagel2003universal}—fit variogram models to encode spatial dependence and can perform well when model assumptions hold~\cite{bamber2013new}. In practice, these methods can be brittle: they rely on fixed correlation structures and Gaussian process assumptions, operate only on sparse bed observations, and rarely leverage ancillary predictors (e.g., velocity fields that correlate with deep troughs). Recent ML methods relax these constraints by learning the spatial operator from data. Kriging Convolutional Networks (KCN)~\cite{appleby2020kriging} combine graph neural network (GNN) propagation with kriging-like inductive bias, while GSI dynamically adapts graph connectivity to capture non-stationary spatial patterns for rainfall interpolation~\cite{li2023rainfall}. Together, these works highlight the value of learned, data-driven spatial priors over hand-crafted ones.

\noindent\textbf{Subglacial bed topography prediction.}
Estimating bed elevation beneath ice sheets is central to mass-balance and sea-level projections. Physics-informed neural networks embed flow physics: for Helheim Glacier, Cheng \etal~\cite{cheng2024forward} impose momentum balance (e.g., shallow-shelf/stream approximations~\cite{macayeal1989large,morland1987unconfined}) as soft constraints to improve thickness inference from sparse data. Broad empirical comparisons by Yi \etal~\cite{yi2023evaluating} find that hybrid schemes (e.g., kriging + ML) perform strongly on Greenland. For Antarctica, DeepBedMap~\cite{leong2020deepbedmap} fuses multi-resolution inputs (surface elevation, velocity) with a GAN to enhance small-scale roughness. Recent learning baselines tailored to topography include DeepTopoNet~\cite{tama2025deeptoponet} and GraphTopoNet~\cite{tama2025improving}, which learn \emph{bed} directly (without physics) and focus on architectural/topological priors rather than leakage-safe validation (see~\autoref{tab:sota}). In contrast, we (i) treat BedMachine as a \emph{prior} and learn \emph{residual} thickness with a DeepLabV3$+$ decoder, (ii) regularize with lightweight, scalable physics (mass conservation and flow-aligned smoothing) plus a non-negativity hinge and radar-confidence weighting, and (iii) evaluate with \emph{spatially aware}, block-wise hold-outs and safety buffers to measure generalization under sparse, regionally shifted observations.

\section{Problem Setup}
\label{sec:setup}

\paragraph{Domain and grid.}
Let $\Omega \subset \mathbb{R}^2$ denote a rectangular region of interest discretized on a uniform grid of size $H \times W$ with pixel spacing $\Delta$ (meters; $\Delta\!\approx\!150$\,m in our experiments). We index grid cells by $(i,j)$ and their coordinates by $(x_i,y_j)$.

\paragraph{Known fields (inputs).}
At each grid cell we assume access to the following geophysical fields:
(i) surface elevation $s(x,y)$ [m];
(ii) surface velocity components $v_x(x,y), v_y(x,y)$ [m\,yr$^{-1}$], collected as $\mathbf{v}(x,y)=(v_x,v_y)$;
(iii) surface mass balance $\mathrm{SMB}(x,y)$ [m\,yr$^{-1}$ ice-equivalent];
(iv) thickness tendency $\partial h/\partial t(x,y)$ (abbrev.\ $\mathrm{\partial h/\partial t}$) [m\,yr$^{-1}$];
(v) a prior bed estimate $b_p(x,y)$ [m] (e.g., BedMachine) and the corresponding prior thickness
\begin{equation}
h_p(x,y) = s(x,y) - b_p(x,y).
\label{eq:hprior}
\end{equation}
We include auxiliary features such as $\nabla s$ (surface gradients) and low-frequency coordinate encodings (e.g., Fourier features) to capture large-scale trends.

\paragraph{Sparse radar observations.}
Let $\mathcal{P}=\{(x_k,y_k,b^{\text{rad}}_k)\}_{k=1}^{N}$ be geolocated radar bed picks within $\Omega$ (sparse and irregular). We define a binary \emph{measurement mask} $m(x,y)\!\in\!\{0,1\}$ indicating the presence of a radar pick on the grid and a confidence map $c(x,y)\!\in\![0,1]$ that decreases with distance to the nearest pick (used later for weighting).

\paragraph{Learning target (residual formulation).}
Rather than regress absolute bed elevation, we predict a \emph{normalized residual thickness} $\hat{r}(x,y)$ with respect to the prior thickness $h_p$:
\begin{equation}
r(x,y) = \sigma_t\,\hat{r}(x,y) + \mu_t,
\label{eq:denorm}
\end{equation}
where $(\mu_t,\sigma_t)$ are robust statistics of residuals computed from radar data on the training region (used only for normalization). The recovered thickness and bed are then
\begin{equation}
\hat{h}(x,y) = h_p(x,y) + r(x,y), \qquad
\hat{b}(x,y) = s(x,y) - \hat{h}(x,y).
\label{eq:reconstruct}
\end{equation}
We enforce $\hat{h}\!\ge\!0$ during training via a non-negativity penalty.

\paragraph{Predictor.}
Let $\Phi(x,y)$ denote the input feature stack at each grid cell (e.g., $\mathrm{SMB}$, $\mathrm{\partial h/\partial t}$, $s$, $v_x$, $v_y$, $\nabla s$, coordinate encodings, and $h_p$). Our network $f_\theta$ maps the full image tensor to the normalized residual field,
\begin{equation}
\hat{r} = f_\theta\big(\Phi\big) \in \mathbb{R}^{H\times W},
\end{equation}
which is de-normalized via \eqref{eq:denorm} and converted to $\hat{b}$ via \eqref{eq:reconstruct}.

\paragraph{Physics residual (used as a regularizer).}
To encourage physical plausibility, we penalize the mass-conservation residual
\begin{equation}
\mathcal{R}(x,y;\hat{h}) \;=\; \frac{\partial \hat{h}}{\partial t} \;+\; \nabla\!\cdot\!\big(\hat{h}\,\mathbf{v}\big) \;-\; \mathrm{SMB},
\label{eq:masscon}
\end{equation}
which should be small in magnitude on average (units m\,yr$^{-1}$). This term is evaluated at multiple spatial scales using smoothed fluxes to reduce discretization noise.

\paragraph{Evaluation regions.}
For spatial generalization, we partition $\Omega$ into train/test blocks either vertically (west/east) or horizontally (south/north). Let $\Omega_{\text{tr}}$ and $\Omega_{\text{te}}$ be the two blocks. We erode each block by a buffer of $\delta$ pixels to obtain \emph{cores} $\mathcal{C}_{\text{tr}}$ and $\mathcal{C}_{\text{te}}$ that are at least $\delta$ pixels from the boundary (preventing receptive-field leakage). All reported metrics are computed on $\mathcal{C}_{\text{te}}$. We also stratify errors by distance to radar using $d_{\text{rad}}(x,y)$, the grid distance to the nearest pick, with bins $[0,2]$, $(2,6]$, and $(6,\infty)$ pixels.

\section{Method}
\label{sec:method}

\begin{figure*}[ht!]
  \centering
  \includegraphics[width=1\textwidth]{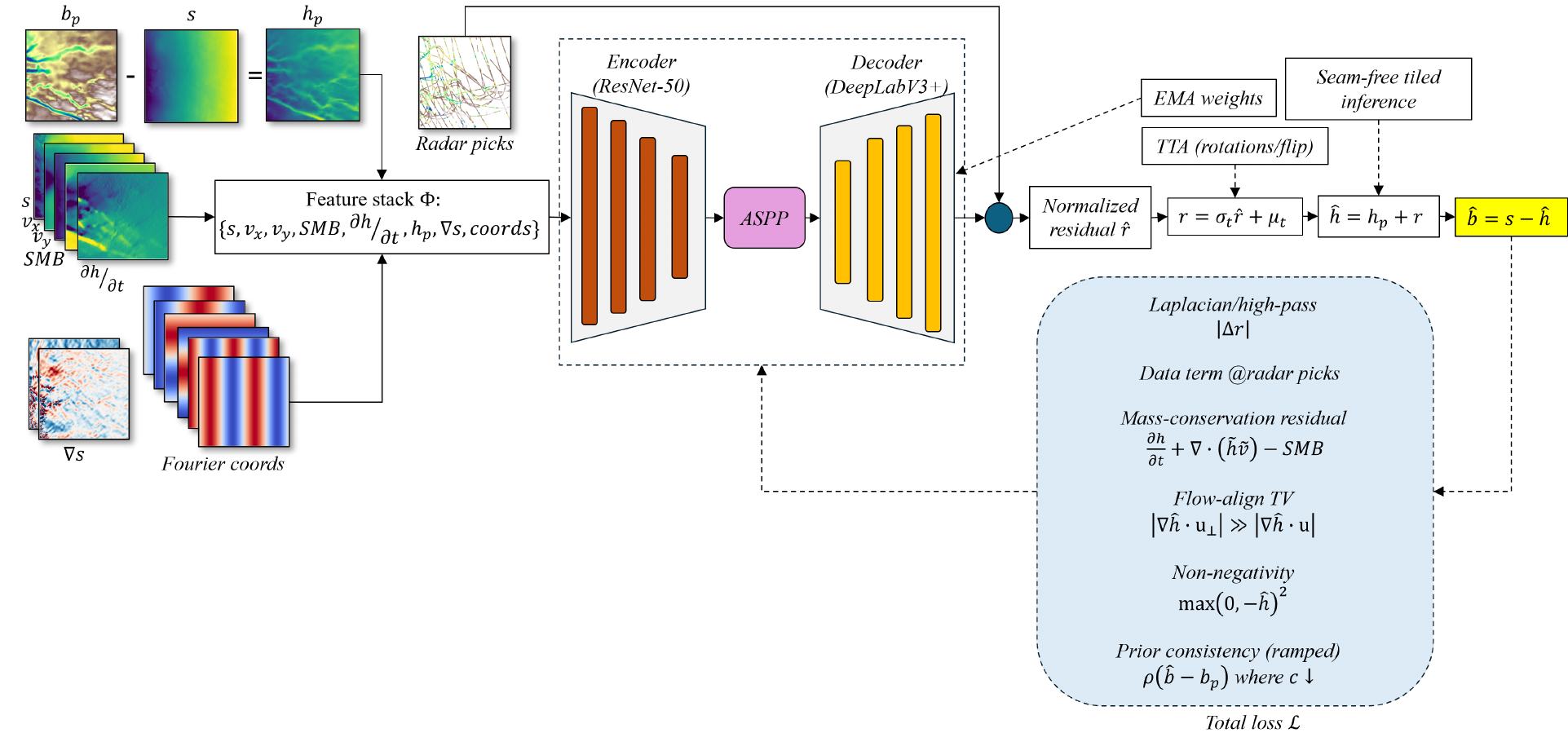}  
  \caption{\textbf{Framework overview.} Inputs (surface, velocity, SMB, thickness tendency) and a BedMachine prior form a feature stack for a DeepLabV3$+$ predictor of normalized residual thickness. De-normalization and reconstruction yield thickness and bed. Training uses data, physics, and regularization losses; inference leverages EMA, test-time augmentation, and seam-free tiling. Evaluation is leakage-safe on held-out cores with distance-to-radar stratification.}
  \label{fig:pipeline}
\end{figure*}

\noindent\textbf{Overview.}
\autoref{fig:pipeline} summarizes our pipeline. Given gridded geophysical fields from \S\ref{sec:setup}—surface $s$, velocity $(v_x,v_y)$, $\mathrm{SMB}$, and $\partial h/\partial t$—together with the BedMachine prior $b_p$, we form the prior thickness $h_p=s-b_p$ and a feature stack $\Phi$ (including $\nabla s$ and low-frequency coordinate encodings). A DeepLabV3$+$ head over a standard encoder (e.g., ResNet-50) predicts a normalized residual thickness $\hat{r}$; we de-normalize to $r=\sigma_t\hat{r}+\mu_t$, reconstruct $\hat{h}=h_p+r$, and obtain the bed $\hat{b}=s-\hat{h}$. Training combines (i) a masked Huber fit to thickness at radar picks, (ii) a multi-scale mass-conservation residual using smoothed fluxes, (iii) flow-aligned total variation (cross-flow $>$ along-flow), (iv) a Laplacian/high-pass penalty on $r$, (v) a non-negativity hinge on $\hat{h}$, and (vi) a ramped prior-consistency term where radar confidence is low. For inference we use EMA weights, 8-way test-time augmentation (rotations/flips with proper vector handling), and seam-free tiled aggregation to produce full-resolution maps.

\subsection{Residual-over-prior predictor}
We predict a \emph{normalized residual thickness} $\hat{r}\in\mathbb{R}^{H\times W}$ with respect to the prior thickness $h_p$ (Sec.~\ref{sec:setup}), and recover thickness and bed via
\[
r = \sigma_t \hat{r} + \mu_t, \qquad
\hat{h} = h_p + r, \qquad
\hat{b} = s - \hat{h}.
\]
The feature stack $\Phi$ at each grid cell includes geophysical fields ($s$, $v_x$, $v_y$, $\mathrm{SMB}$, $\mathrm{\partial h/\partial t}$), auxiliary cues (e.g., $\nabla s$), low-frequency coordinate encodings, and $h_p$.

We instantiate $f_\theta$ as a DeepLabV3$+$ decoder fed by a standard vision encoder (ResNet-50 with output stride 16). The encoder produces a high-level map and a low-level map; the decoder fuses atrous spatial pyramid pooling (ASPP) features with low-level features through a $1\times1$ projection, then upsamples to full resolution. We adopt group normalization and leaky ReLU in all convolutional blocks for stability on varied batch sizes. All scalar channels are normalized per-channel; vector channels ($v_x,v_y$) receive geometry-aware augmentations (Sec.~\ref{sec:inference}).

\subsection{Training objective}
\label{sec:loss}
The total loss is a weighted sum
\begin{multline}\label{eq:total_loss_method}
\mathcal{L} =
\lambda_{\text{data}}\,\mathcal{L}_{\text{radar}}
+ \lambda_{\text{phys}}\,\mathcal{L}_{\text{mass}}
+ \lambda_{\text{tv}}\,\mathcal{L}_{\text{flowTV}} \\
+ \lambda_{\text{lap}}\,\mathcal{L}_{\text{lap}}
+ \lambda_{\ge 0}\,\mathcal{L}_{\ge 0}
+ \lambda_{\text{prior}}\,\mathcal{L}_{\text{prior}} \, .
\end{multline}
We use a robust Huber penalty $\rho_\delta(t)$ for all residuals unless stated.

\paragraph{Data term at radar picks.}
Let $m(x,y)\in\{0,1\}$ indicate a radar pick on the grid and $c(x,y)\in[0,1]$ be a confidence that decays with distance to the nearest pick (Sec.~\ref{sec:setup}). We compare \emph{thickness} at picks to avoid subtracting two noisy beds:
\begin{equation}
\mathcal{L}_{\text{radar}}
= \frac{1}{Z}\!\sum_{(x,y)} m(x,y)\, w(x,y)\,\rho_{\delta}\!\big(\hat{h}(x,y)-h^{\text{rad}}(x,y)\big),
\end{equation}
with weights $w(x,y)=\max(\epsilon,\,c(x,y))$ and $Z=\sum m\,w$. In practice, $h^{\text{rad}}$ is obtained by splatting radar bed picks to grid and subtracting $s$. We set $c(x,y)=\text{exp}(-d(x,y)/\tau)$, where $d$ is distance to the nearest radar pick and $\tau$ a decay scale; physics/prior terms use $(1-c)$ or $(1-c)^{2}$ to emphasize radar-sparse areas.

\paragraph{Mass-conservation regularizer.}
We penalize the multi-scale residual of the thickness continuity (units m\,yr$^{-1}$):
\begin{equation}
\mathcal{L}_{\text{mass}}
= \sum_{k\in\mathcal{S}} \alpha_k \;\mathbb{E}\Big[\, \rho_{\delta}
\big( \mathrm{\partial h \partial t}_k + \nabla\!\cdot(\tilde{h}_k \tilde{\mathbf{v}}_k) - \mathrm{SMB}_k \big)\,\Big].
\label{eq:massloss}
\end{equation}
Here $\mathcal{S}=\{1,2,4\}$ denotes average-pooling scales; $(\cdot)_k$ indicates downsampling by $k$; and tildes denote Gaussian-smoothed fields to mitigate discretization noise in the divergence. We weight the penalty more strongly away from radar by multiplying inside the expectation with $(1-c)^q$ (typically $q\in[1,2]$).

\paragraph{Flow-aligned total variation.}
Let $\mathbf{u}=\mathbf{v}/(\|\mathbf{v}\|+\epsilon)$ be the unit flow direction and $\mathbf{u}_\perp$ be an orthogonal unit vector. We discourage cross-flow gradients more than along-flow gradients:
\begin{equation}
\mathcal{L}_{\text{flowTV}}
= \mathbb{E}\!\big[\, \beta_\perp \,| \nabla \hat{h} \cdot \mathbf{u}_\perp | + \beta_\parallel \,| \nabla \hat{h} \cdot \mathbf{u} | \,\big], 
\quad \beta_\perp > \beta_\parallel .
\end{equation}

\paragraph{Laplacian / high-pass damping.}
To suppress ringing artifacts without oversmoothing valleys, we penalize the discrete Laplacian of the \emph{residual} field:
\begin{equation}
\mathcal{L}_{\text{lap}} \;=\; \mathbb{E}\!\big[\, |\Delta r| \,\big].
\end{equation}

\paragraph{Non-negativity of thickness.}
We softly enforce $\hat{h}\ge 0$ using a hinge:
\begin{equation}
\mathcal{L}_{\ge 0} \;=\; \mathbb{E}\!\big[\, \max(0,\,-\hat{h})^2 \,\big].
\end{equation}

\paragraph{Prior consistency ramp.}
Where radar confidence is low, we nudge the reconstruction toward the prior bed, but avoid fighting the data near picks:
\begin{equation}
\mathcal{L}_{\text{prior}}
= \mathbb{E}\Big[\, (1-m)\,(1-c)^2 \;\rho_{\delta_p}\big(\hat{b}-b_p\big) \Big].
\end{equation}
We \emph{ramp} $\lambda_{\text{prior}}$ over training (small early, larger later) so the network first fits data and physics before relying on the prior.

\paragraph{Weight schedules.}
We linearly ramp $\lambda_{\text{phys}}$ from $0$ to its target over most of training (e.g., first $90\%$ of epochs) to avoid early over-regularization, and ramp $\lambda_{\text{prior}}$ between mid and late training. Fixed small $\lambda_{\text{lap}}$ and $\lambda_{\ge 0}$ stabilize optimization.

\subsection{Inference: TTA, seam-free tiling, and EMA}
\label{sec:inference}
\paragraph{Exponential moving average (EMA).}
We maintain $\theta_{\text{ema}}\leftarrow \alpha\,\theta_{\text{ema}} + (1-\alpha)\,\theta$ (with $\alpha\approx 0.999$) and use $f_{\theta_{\text{ema}}}$ for validation/testing, which improves calibration and reduces flicker across tiles.

\paragraph{Test-time augmentation (TTA).}
We apply 8 geometric transforms (4 rotations $\times$ horizontal flip) to inputs, invert them on outputs, and average predictions. Vector channels $(v_x,v_y)$ and derived gradient channels are transformed with the correct rotation/flip rules to preserve physical meaning.

\paragraph{Seam-free tiled inference.}
For large grids we use overlapping $P\times P$ tiles but only accumulate the \emph{core} region (excluding a $b$-pixel border) from each tile into the output canvas, averaging overlaps. Reflect padding at tile boundaries avoids edge artifacts. This yields a seamless full-resolution prediction while keeping memory bounded.

\paragraph{Leakage-safe evaluation.}
During training and evaluation, the block-wise splits and safety buffers from Sec.~\ref{sec:setup} ensure that receptive fields of training tiles do not touch the test core. All reported metrics are computed strictly on the held-out core.

\section{Experimental Protocol}
\label{sec:protocol}

\subsection{Datasets and Splits}
\begin{figure}[ht!]
    \centering
    \includegraphics[width=0.3\textwidth]{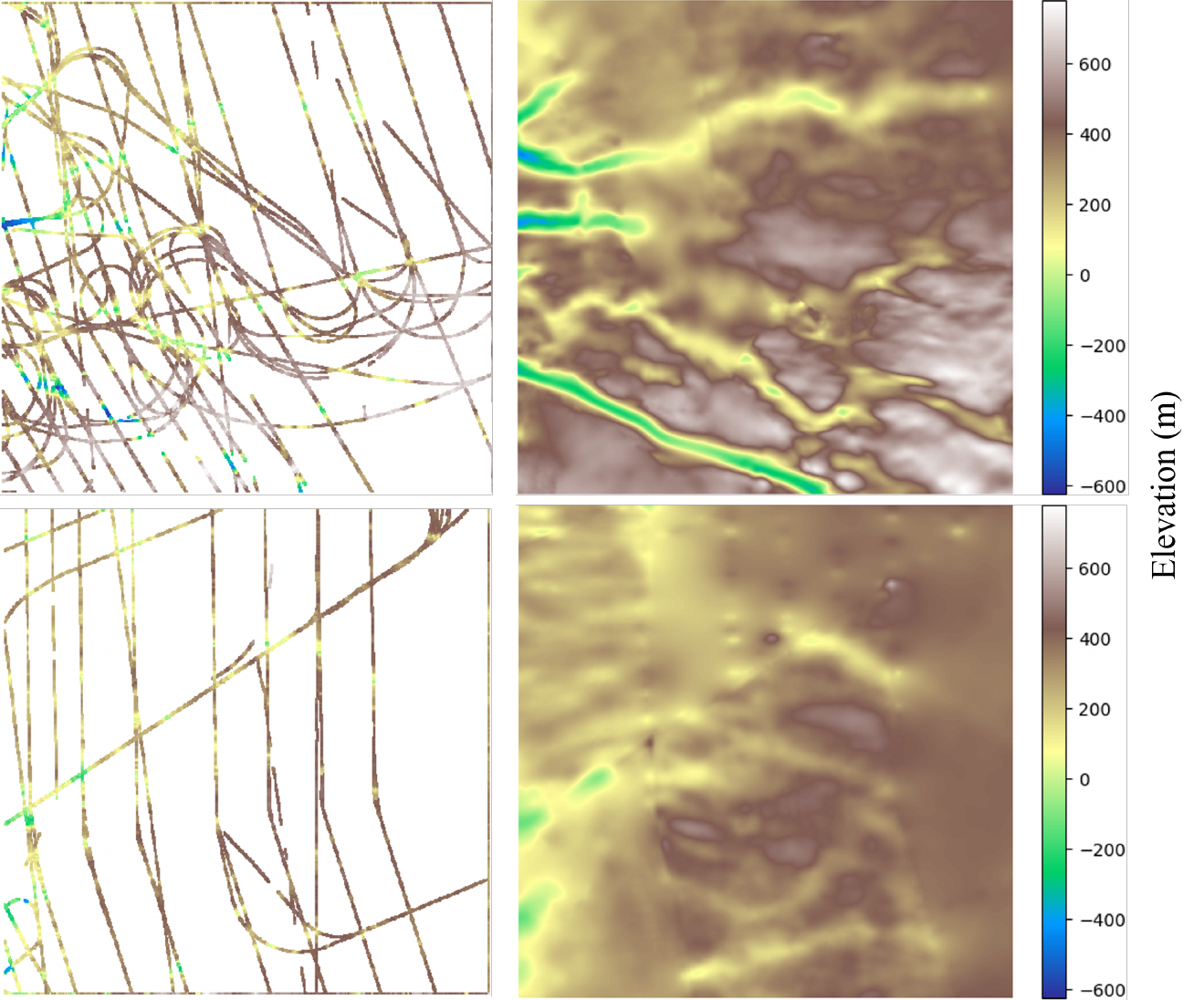}
    \caption{\textbf{Radar picks and BedMachine prior.} Left: geolocated radar bed picks (colored by observed bed elevation). Right: prior bed $b_p$ (BedMachine) on the same grid. Rows show Sub-region I (top) and Sub-region II (bottom) of Upernavik Isstr{\o}m.}
    \label{dataset}
\end{figure}

We evaluate on two Greenland sub-regions ((i.e., Upernavik Isstr{\o}m)) discretized on uniform grids of size $H\times W$ (e.g., $600\times 600$). All rasters are in EPSG:3413 (NSIDC Sea Ice Polar Stereographic North), resampled to $\Delta \approx 150\,\mathrm{m}$. Inputs comprise the geophysical fields from Sec.~\ref{sec:setup} (surface elevation $s$, velocity $v_x,v_y$, surface mass balance $\mathrm{SMB}$, and thickness tendency $\partial h/\partial t$), a prior bed $b_p$ (BedMachine), and sparse radar picks. \autoref{dataset} illustrates, for each sub-region, the spatial distribution of radar pick locations (left) and the corresponding prior bed $b_p$ resampled to our grid (right); the pronounced sparsity and anisotropy of the picks motivate our residual-over-prior design used in evaluation. 

To assess spatial generalization without receptive-field leakage, we adopt two orthogonal \emph{block-wise} splits:
(i) a \textbf{vertical} split (west train / east test) defined by the median $x$-coordinate; and
(ii) a \textbf{horizontal} split (south train / north test) defined by the median $y$-coordinate.
We erode each block by a safety buffer of $\delta=96\,\mathrm{px}$ to obtain \emph{cores} $\mathcal{C}_{\text{tr}}$ and $\mathcal{C}_{\text{te}}$ that are at least $\delta$ pixels from the train/test boundary. All models are trained using tiles whose \emph{core} region lies fully inside $\mathcal{C}_{\text{tr}}$, and all reported metrics are computed strictly on $\mathcal{C}_{\text{te}}$.

\subsection{Baselines}
We compare against strong classical and deep baselines:
\begin{itemize}
\item \textit{Inverse Distance Weighting (IDW)}. We use classical IDW~\cite{shepard1968two} with a $k$-NN moving neighborhood and power $p$, including exact-hit handling; we report grid-sampled radar errors and optionally perform leave-one-out cross-validation following geostatistical practice, and we contextualize IDW choices using a recent comparative review~\cite{li2014spatial}. 
    \item \textit{Kriging}. We implement residual kriging—kriging the bed residuals relative to a BedMachine prior and adding the prior back—using simple kriging (known mean 0 for residuals) or local ordinary kriging with a $k$-NN moving neighborhood. We estimate a robust empirical semivariogram and fit an exponential model via weighted least squares~\cite{cressie1985fitting,Hengl2007RegressionKriging,cressie1980robust}.
    \item \textit{Convolutional Neural Network (CNN)}. Our CNN baseline is a fully convolutional residual network with dilated (atrous) convolutions for large receptive fields~\cite{he2016deep,Yu2016Dilated,yu2017dilated}, Group Normalization for batch-size robustness~\cite{wu2018group}, and and Leaky ReLU activations.
    \item \textit{U-Net}. Our model is a UNet-style encoder–decoder with skip connections~\cite{ronneberger2015u}, augmented by an ASPP bottleneck for multi-scale context~\cite{chen2017rethinking,chen2018encoder,Yu2016Dilated} and and Group Normalization for small-batch stability~\cite{wu2018group}, using Leaky ReLU activations.
    \item \textit{Attention U-Net}. We adopt an Attention U-Net architecture~\cite{ronneberger2015u,oktay2018attention}, with Group Normalization~\cite{wu2018group} and Leaky ReLU activations, using bilinear upsampling to avoid checkerboard artifacts~\cite{odena2016deconvolution}.
    \item \textit{Feature Pyramid Network (FPN)}. We use a ResNet-50 + FPN decoder~\cite{lin2017feature}, with Group Normalization for small-batch stability~\cite{wu2018group}, and and a semantic-FPN head that upsamples and concatenates multi-level features for dense regression~\cite{kirillov2019panoptic}.
\end{itemize}

All learning baselines are trained on the same $\mathcal{C}_{\text{tr}}$ tiles with identical optimization, augmentation, and early-stopping criteria for fairness. 

\begin{table*}[ht!]
\centering
\caption{\textbf{Main quantitative comparison on held-out test cores.}
Two sub-regions evaluated under vertical and horizontal block-wise splits. “BedMachine” rows report agreement with the BedMachine raster. “Radar” rows report errors at held-out radar picks. Best/2$^{\text{nd}}$/3$^{\text{rd}}$ per column are shaded with a three-level tint. Distance-stratified RMSE appears in the Supplement~\autoref{sec:disradar}.}
\label{tab:main_dl}
\begin{adjustbox}{width=\textwidth}
\begin{tabular}{l|c|cccccc|cccccc}
\hline
\multirow{3}{*}{Method} & \multirow{3}{*}{Reference Data} & \multicolumn{6}{c|}{Sub-Region I}                              & \multicolumn{6}{c}{Sub-Region II}                             \\\cline{3-14}
                       &                                 & \multicolumn{3}{c}{Vertical} & \multicolumn{3}{c|}{Horizontal} & \multicolumn{3}{c}{Vertical} & \multicolumn{3}{c}{Horizontal} \\\cline{3-14}
&& \multicolumn{1}{c}{MAE$\downarrow$} & \multicolumn{1}{c}{RMSE$\downarrow$} & \multicolumn{1}{c}{{R$^{2}\uparrow$}} & \multicolumn{1}{c}{MAE$\downarrow$} & \multicolumn{1}{c}{RMSE$\downarrow$} & \multicolumn{1}{c|}{R$^{2}\uparrow$} & \multicolumn{1}{c}{MAE$\downarrow$} & \multicolumn{1}{c}{RMSE$\downarrow$} & \multicolumn{1}{c}{R$^{2}\uparrow$} & \multicolumn{1}{c}{MAE$\downarrow$} & \multicolumn{1}{c}{RMSE$\downarrow$} & \multicolumn{1}{c}{{R$^{2}\uparrow$}} \\
\hline
\multirow{2}{*}{IDW~\cite{shepard1968two}}   &BedMachine&345.84&379.53&-7.980&184.95&218.49&-2.131&308.39&321.63&-51.354&183.38&208.61&-5.145\\
& Radar&290.18&318.20&-5.304&226.24&251.44&-2.029&302.48&327.32&-5.671&184.35&214.47&-2.166\\\hline
\multirow{2}{*}{Kriging~\cite{cressie1985fitting,Hengl2007RegressionKriging,cressie1980robust}}   &BedMachine &22.47&28.03&0.951&13.23&27.89&0.78&74.56&87.98&-2.918&40.79&55.33&0.568\\
& Radar&90.95&117.97&0.134&94.01&126.65&0.231&\rdfirst{87.71}&\rdfirst{123.89}&\rdfirst{0.044}&81.15&107.93&0.198 \\\hline
\multirow{2}{*}{CNN~\cite{he2016deep,wu2018group,yu2017dilated}}   & BedMachine & 18.4     & 29.4      & 0.946 & 13.88     & 23.2& 0.965& 6.69& 10.5& 0.944  & 5.99      & \bmsecond{9.45}      & \bmsecond{0.987}  \\
& Radar& 87.2     & 112.05    & 0.218 & \rdthird{93.3}      & \rdsecond{125}       & \rdsecond{0.251}  & 99.72    & 129.91   & -0.051 & \rdthird{59.42}     & \rdsecond{88.74}     & \rdsecond{0.458} \\\hline
\multirow{2}{*}{U-Net~\cite{ronneberger2015u}}&BedMachine&\bmsecond{13.68}&	\bmthird{24.66}&\bmthird{0.962}&	\bmsecond{10.19}&	\bmsecond{18.32}&	\bmsecond{0.978}&	\bmsecond{3.83}&	\bmsecond{6.92}&	\bmsecond{0.976}&	\bmsecond{5.58}&	10.91&	0.983\\
&Radar&\rdsecond{75.6}&	\rdsecond{99.76}&	\rdsecond{0.38}&	94.03&	126.66&	0.231&	97.24&	126.02&	0.011&	60.23&	89.83&	0.445\\\hline
\multirow{2}{*}{Att. U-Net~\cite{ronneberger2015u,oktay2018attention}}&BedMachine&\bmthird{13.92}&	\bmsecond{24.02}&	\bmsecond{0.964}&	\bmthird{11.84}&	\bmthird{20.75}&	\bmthird{0.972}&	\bmthird{3.93}&	\bmthird{8.11}&	\bmthird{0.967}&	\bmthird{5.7}&	12.32&	0.979\\
&Radar&84.5&	111.1&	0.231&	94.16&	126.17&	0.237&	\rdsecond{96.33}&	\rdsecond{124.87}&	\rdsecond{0.029}&	\rdfirst{58.78}&	\rdfirst{88.36}&\rdfirst{0.463}\\\hline
\multirow{2}{*}{FPN~\cite{lin2017feature}}&BedMachine &26.78&	35.93&	0.92&	17.35&	26.91&	0.953&	7.75&	11.25&	0.936&	6.66&	\bmthird{10.18}&	\bmthird{0.985}\\
&Radar&\rdfirst{63.22}&	\rdfirst{80.65}&\rdfirst{0.595}&\rdfirst{93}&\rdfirst{124.4}&\rdfirst{0.258}&	\rdthird{96.62}&	\rdthird{125.49}&	\rdthird{0.019}&	\rdsecond{59.16}&	\rdthird{88.98}&	\rdthird{0.455}\\\hline
\multirow{2}{*}{\textbf{Ours}}&BedMachine &\bmfirst{7.59}&	\bmfirst{10.54}&	\bmfirst{0.993}&	\bmfirst{4.98}&	\bmfirst{8.12}&	\bmfirst{0.996}&	\bmfirst{2.05}&	\bmfirst{3.33}&	\bmfirst{0.994}&	\bmfirst{1.96}&	\bmfirst{3.05}&	\bmfirst{0.999}\\
&Radar&\rdthird{80.66}&	\rdthird{105.04}&	\rdthird{0.313}&	\rdsecond{93.15}&	\rdthird{125.17}&	\rdthird{0.249}&	97.3&	125.94&	0.012&	60.16&	89.81&	0.445\\
\hline
\end{tabular}
\end{adjustbox}
\end{table*}

\begin{table*}[ht!]
\centering
\caption{\textbf{Structural and perceptual metrics on held-out test cores.}
SSIM$\uparrow$, PSNR$\uparrow$ (dB), and $|\Delta\mathrm{TRI}|\downarrow$ (mean absolute difference in Terrain Ruggedness Index) for two Greenland sub-regions under vertical and horizontal block-wise splits. Best/2$^{\text{nd}}$/3$^{\text{rd}}$ per column are shaded with a three-level tint.}

\label{tab:main_dl2}
\begin{adjustbox}{width=\textwidth}
\begin{tabular}{l|cccccc|cccccc}
\hline
\multirow{3}{*}{Method}& \multicolumn{6}{c|}{Sub-Region I}& \multicolumn{6}{c}{Sub-Region II}\\\cline{2-13}
& \multicolumn{3}{c}{Vertical}& \multicolumn{3}{c|}{Horizontal}& \multicolumn{3}{c}{Vertical}& \multicolumn{3}{c}{Horizontal}\\\cline{2-13}
& SSIM$\uparrow$                 & \multicolumn{1}{c}{PSNR$\uparrow$} & \multicolumn{1}{c}{$\mid\triangle \text{TRI}\mid\downarrow$} & \multicolumn{1}{c}{SSIM$\uparrow$} & \multicolumn{1}{c}{PSNR$\uparrow$} & \multicolumn{1}{c|}{$\mid\triangle \text{TRI}\mid\downarrow$} & SSIM$\uparrow$                 & \multicolumn{1}{c}{PSNR$\uparrow$} & \multicolumn{1}{c}{$\mid\triangle \text{TRI}\mid\downarrow$} & \multicolumn{1}{c}{SSIM$\uparrow$} & \multicolumn{1}{c}{PSNR$\uparrow$} & \multicolumn{1}{c}{$\mid\triangle \text{TRI}\mid\downarrow$} \\
\hline
IDW~\cite{shepard1968two}   &0.323&6.54&8.83&0.544&13.24&7.04&0.247&4.01&3.79&0.569&10.22&8.85\\
Kriging~\cite{cressie1985fitting,Hengl2007RegressionKriging,cressie1980robust}   &\bmsecond{0.994}&37.70&\bmsecond{2.71}&\bmsecond{0.993}&39.98&\bmsecond{2.2}&0.966&23.73&3.76&0.982&27.76&\bmsecond{1.49}\\
CNN~\cite{he2016deep,wu2018group,yu2017dilated}   &0.981&	37.29&	\bmthird{5.46}&	0.981&	39.35&	\bmthird{6.28}&	0.992&	42.19&	3.62&	\bmsecond{0.993}&	\bmsecond{43.11}&	\bmthird{2.97}\\
U-Net~\cite{ronneberger2015u}&\bmthird{0.983}&	\bmthird{38.82}&	9.53&	\bmthird{0.984}&	\bmsecond{41.4}&	6.55&	\bmsecond{0.995}&	\bmsecond{45.82}&	\bmsecond{2.01}&	0.99&	41.86&	3.01\\
Att. U-Net~\cite{ronneberger2015u,oktay2018attention}&0.982&	\bmsecond{39.04}&	9.77&	0.981&	\bmthird{40.32}&	7.67&	\bmthird{0.994}&	\bmthird{44.44}&	\bmthird{2.24}&	0.988&	40.81&	3.62\\
FPN~\cite{lin2017feature}&0.972&	35.55&	13.72&	0.971&	38.06&	9.86&	0.985&	41.57&	5.06&	\bmthird{0.991}&	\bmthird{42.46}&	3.5\\
\textbf{Ours}&\bmfirst{0.998}&	\bmfirst{46.2}&	\bmfirst{2.53}&	\bmfirst{0.998}&	\bmfirst{48.46}&	\bmfirst{1.44}&	\bmfirst{0.999}&\bmfirst{52.17}&\bmfirst{0.66}&\bmfirst{0.999}&\bmfirst{52.94}&	\bmfirst{0.66}\\
\hline
\end{tabular}
\end{adjustbox}
\end{table*}

\begin{table*}[ht!]
\centering
\caption{\textbf{Ablations of loss components on held-out test cores.}
Effect of removing mass-conservation (\(\mathcal{L}_{\text{mass}}\)), prior-consistency (\(\mathcal{L}_{\text{prior}}\)), non-negativity, and radar data fit (\(\mathcal{L}_{\text{radar}}\)). 
}

\label{tab:ablation}
\begin{adjustbox}{width=\textwidth}
\begin{tabular}{l|c|cccccc|cccccc}
\hline
\multirow{3}{*}{Method} & \multirow{3}{*}{Reference Data} & \multicolumn{6}{c|}{Sub-Region I}                              & \multicolumn{6}{c}{Sub-Region II}                             \\\cline{3-14}
                       &                                 & \multicolumn{3}{c}{Vertical} & \multicolumn{3}{c|}{Horizontal} & \multicolumn{3}{c}{Vertical} & \multicolumn{3}{c}{Horizontal} \\\cline{3-14}
&& \multicolumn{1}{c}{MAE$\downarrow$} & \multicolumn{1}{c}{RMSE$\downarrow$} & \multicolumn{1}{c}{{R$^{2}\uparrow$}} & \multicolumn{1}{c}{MAE$\downarrow$} & \multicolumn{1}{c}{RMSE$\downarrow$} & \multicolumn{1}{c|}{R$^{2}\uparrow$} & \multicolumn{1}{c}{MAE$\downarrow$} & \multicolumn{1}{c}{RMSE$\downarrow$} & \multicolumn{1}{c}{R$^{2}\uparrow$} & \multicolumn{1}{c}{MAE$\downarrow$} & \multicolumn{1}{c}{RMSE$\downarrow$} & \multicolumn{1}{c}{{R$^{2}\uparrow$}} \\
\hline
\multirow{2}{*}{w/o $\mathcal{L}_{mass}$}&BedMachine&8.55&	11.44&	0.994&	5.66&	10.69&	0.993&	2.69&	4.68&	0.989&	8.8&	16.1&	0.99\\
& Radar&81.31&	106.22&	0.298&	93.66&	125.86&	0.241&	95.45&	124.12&	0.041&	59.76&	89.19&	0.452
\\\hline
\multirow{2}{*}{w/o $\mathcal{L}_{prior}$}&BedMachine&24.02&	31.25&	0.939&	27.54&	38.14&	0.905&	16.83&	21.94&	0.756&	21.12&	24.86&	0.913\\
& Radar&93.74&	122.03&	0.073&	96.77&	129.21&	0.2&	92.4&	120.63&	0.094&	65.45&	98.41&	0.333\\\hline
\multirow{2}{*}{w/o non-negativity}&BedMachine&430.64&	442.22&	-11.191&	95.45&	111.61&	0.183&	293.95&	298.65&	-44.14&	80.22&	95.14&	-0.278\\
& Radar&558.49&	581.98&	-20.089&	156.4&	191.08&	-0.749&	326.09&	357.3&	-6.949&	110.65&	114.53&	-0.438\\\hline
\multirow{2}{*}{w/o $\mathcal{L}_{radar}$}&BedMachine&10.21&	15.29&	0.985&	5.34&	9.06&	0.995&2.79&4.50&0.990&2.32&3.99&0.998\\
& Radar&82.96&	108.59&	0.266&	94.28&	127&	0.227&96.10&124.33&0.037&60.19&89.84&0.444\\\hline
\end{tabular}
\end{adjustbox}
\end{table*}

\subsection{Metrics}

All metrics are computed on the held-out test core $\mathcal{C}_{\text{te}}$ to avoid receptive-field leakage as follows.
\begin{itemize}
    \item \textit{Pixel-wise agreement.} We report MAE, RMSE, and $R^2$ between the predicted bed $\hat{b}$ and the reference on $\mathcal{C}_{\text{te}}$.
    \item \textit{Perceptual/structural similarity.} We report SSIM (range $[0,1]$, higher is better) and PSNR (dB, higher is better) between $\hat{b}$ and the reference on $\mathcal{C}_{\text{te}}$. PSNR is computed as $10\log_{10}(R^2/\mathrm{MSE})$ with $R$ taken as the dynamic range of the reference (max$-$min) within the evaluation region.
    \item \textit{Morphology (roughness consistency).} To assess whether bed \emph{texture} is preserved, we compute the Terrain Ruggedness Index (TRI)~\cite{riley1999index} on both maps using a $3\times3$ neighborhood and report the mean absolute difference $|\Delta\mathrm{TRI}|$ on $\mathcal{C}_{\text{te}}$. This summarizes roughness/valley structure beyond pixel-wise error. 
    \item \textit{Radar-only errors}. Independently of image-wide metrics, we sample $\hat{b}$ at held-out radar pick locations inside $\mathcal{C}_{\text{te}}$ and report MAE and RMSE against the pick values to quantify fidelity at observations.
    \item \textit{Distance-to-radar stratification.} For completeness, we also report RMSE stratified by distance to the nearest radar pick in three bins ($0$–$2$, $2$–$6$, $>6$ px).
\end{itemize}

\subsection{Implementation Details}
Unless noted, we use patch size $P=256$, stride $64$, and a core border of $b=96\,\mathrm{px}$ for leakage-safe tiling. Training uses AdamW with cosine restarts, test-time augmentation (8-way rotations/flips with vector-aware transforms), EMA weights for validation/testing, and seam-free tiled inference. Physics and prior-consistency weights are linearly ramped (physics over the first $\sim\!90\%$ of epochs; prior from mid to late training). Unless noted: $\lambda_{\text{data}}=2.0$, $\lambda_{\text{phys}}=10^{-2}$ (linear ramp 0$\to$target over the first 90\% of epochs), $\lambda_{\text{tv}}=5\!\times\!10^{-4}$, $\lambda_{\text{lap}}=2\!\times\!10^{-4}$, $\lambda_{\ge 0}=10^{-3}$, and $\lambda_{\text{prior}}=5\!\times\!10^{-3}$ (ramp from epoch 30\% to 90\%). Confidence weighting uses $c$ for $\mathcal{L}_{\text{radar}}$, $(1-c)$ for $\mathcal{L}_{\text{mass}}$, $(1-c)^2$ for $\mathcal{L}_{\text{prior}}$; we mask $\mathcal{L}_{\text{prior}}$ at picks and attenuate it on steep slopes. Huber $\delta$: 1.0 for $\mathcal{L}_{\text{radar}}$, 5.0 for $\mathcal{L}_{\text{mass}}$, 10.0 for $\mathcal{L}_{\text{prior}}$.

Optimization: AdamW (learning rate $1\!\times\!10^{-4}$, weight decay $1\!\times\!10^{-4}$) with cosine warm restarts (T$_0{=}500$, T$_\text{mult}{=}2$); training runs up to 6000 epochs with early stopping (patience 2000 epochs) monitored on masked radar-thickness fit inside the train core; EMA decay $0.999$; best checkpoint selected by validation RMSE. Further low-level settings are given in the Supplement~\autoref{sup:impldet}. 

\begin{table*}[ht!]
\caption{\textbf{Qualitative comparison on held-out test cores for Sub-region I.}
For each method (rows) and split (columns: \emph{Vertical}, \emph{Horizontal}), we show triplets:
\emph{Prediction} $\mid$ \emph{Prior} $b_p$ (BedMachine) $\mid$ \emph{Difference} (BedMachine $-\ \hat{b}$). All panels use consistent color limits; difference maps use a zero-centered diverging colormap (white $\approx$ 0 m).}
\label{tab:map}
\centering
\begin{tabular}{l|l}
\hline
Vertical&Horizontal\\ \hline
\multicolumn{2}{c}{Prediction $\mid$ Prior $b_{p}$ $\mid$ Difference}\\\hline
IDW~\cite{shepard1968two}\\
\begin{minipage}[c]{0.36\textwidth}\includegraphics[width=1\textwidth]{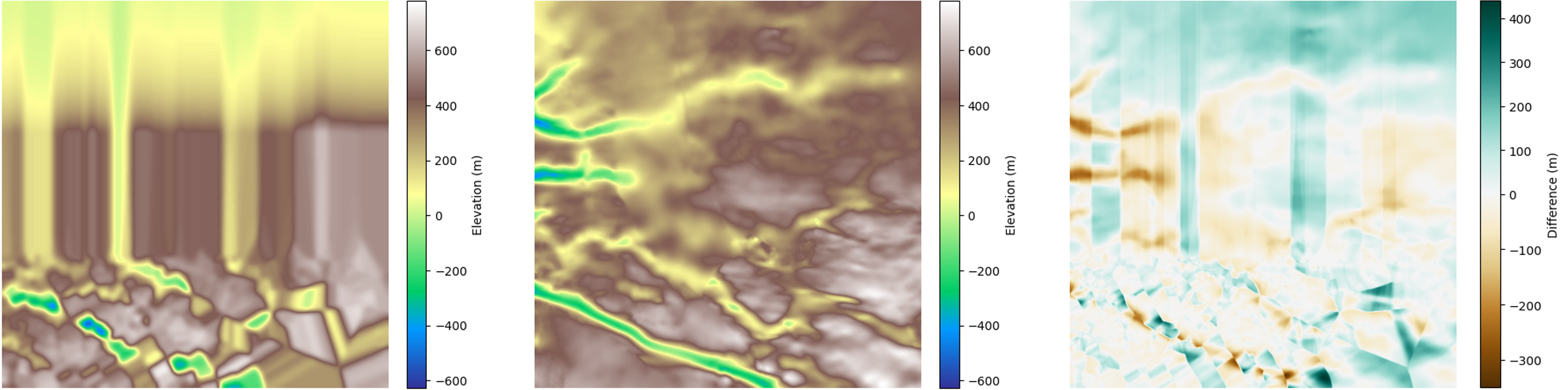}\end{minipage}&\begin{minipage}[c]{0.36\textwidth}\includegraphics[width=1\textwidth]{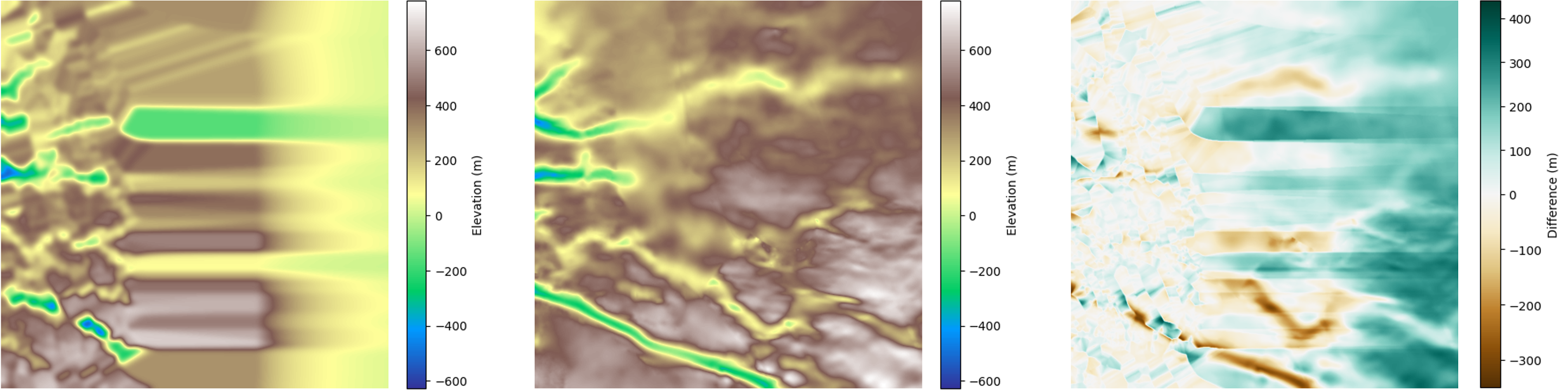}\end{minipage}\\
Kriging~\cite{cressie1985fitting,Hengl2007RegressionKriging,cressie1980robust}\\
\begin{minipage}[c]{0.36\textwidth}\includegraphics[width=1\textwidth]{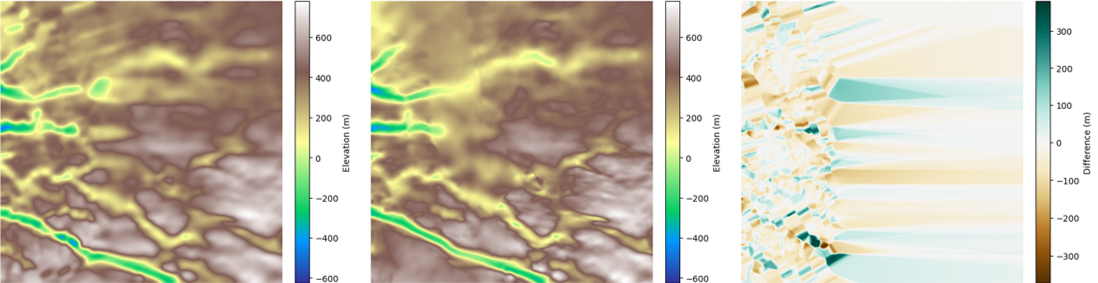}\end{minipage}&\begin{minipage}[c]{0.36\textwidth}\includegraphics[width=1\textwidth]{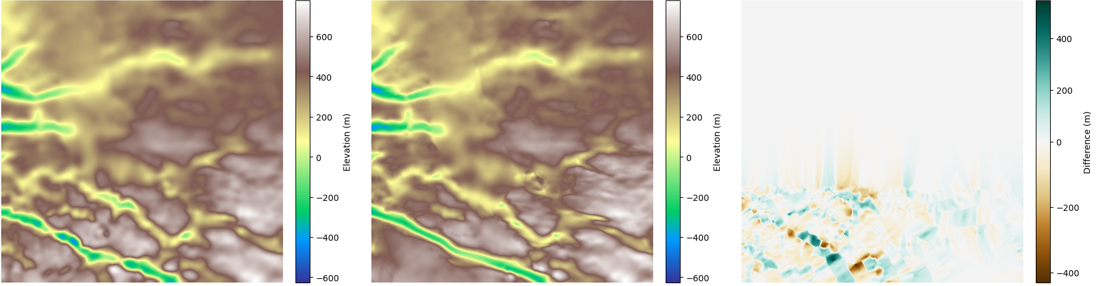}\end{minipage}\\
CNN~\cite{he2016deep,wu2018group,yu2017dilated}\\
\begin{minipage}[c]{0.36\textwidth}\includegraphics[width=1\textwidth]{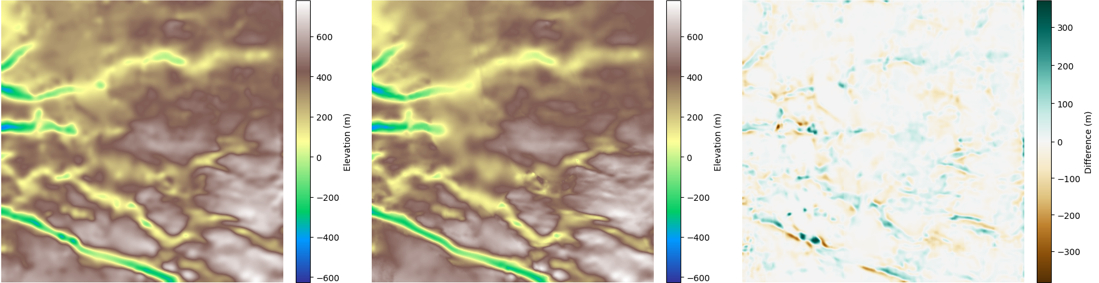}\end{minipage}&\begin{minipage}[c]{0.36\textwidth}\includegraphics[width=1\textwidth]{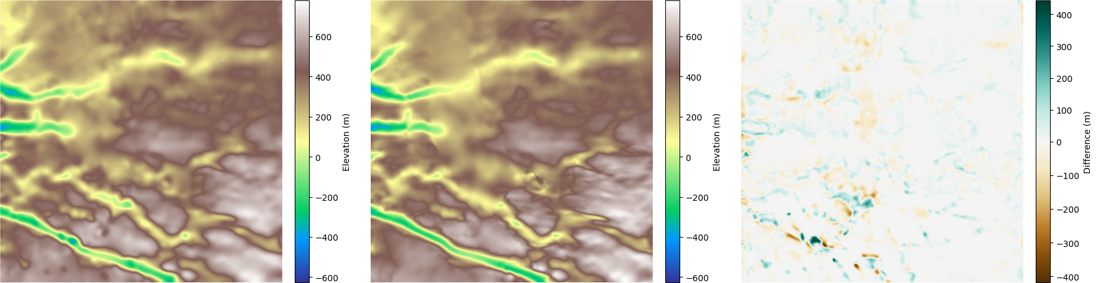}\end{minipage}\\
U-Net~\cite{ronneberger2015u}\\
\begin{minipage}[c]{0.36\textwidth}\includegraphics[width=1\textwidth]{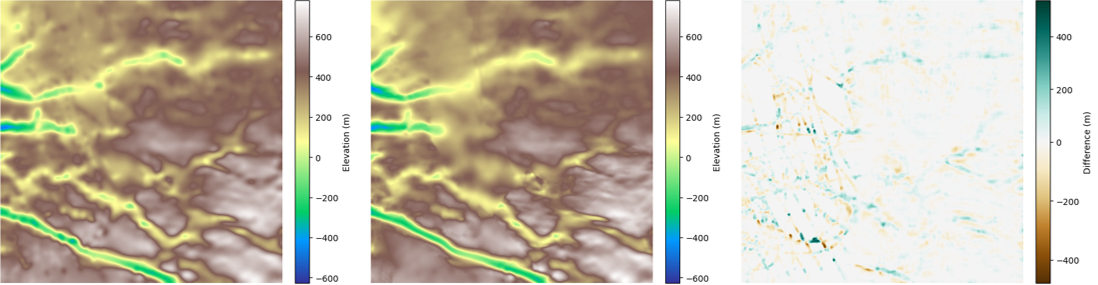}\end{minipage}&\begin{minipage}[c]{0.36\textwidth}\includegraphics[width=1\textwidth]{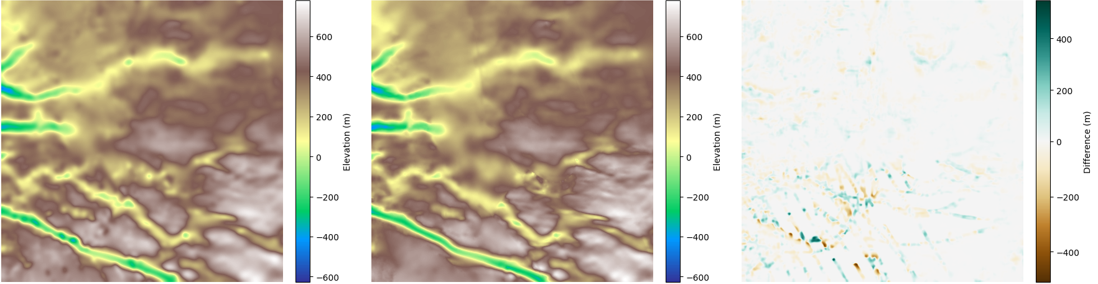}\end{minipage}\\
Att. U-Net~\cite{ronneberger2015u,oktay2018attention}\\
\begin{minipage}[c]{0.36\textwidth}\includegraphics[width=1\textwidth]{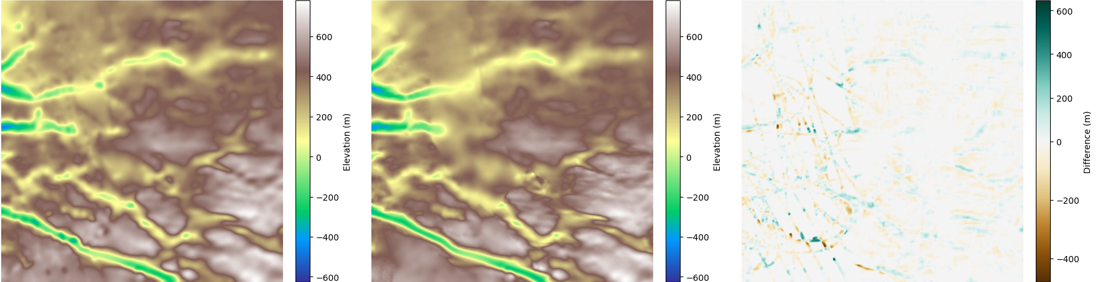}\end{minipage}&\begin{minipage}[c]{0.36\textwidth}\includegraphics[width=1\textwidth]{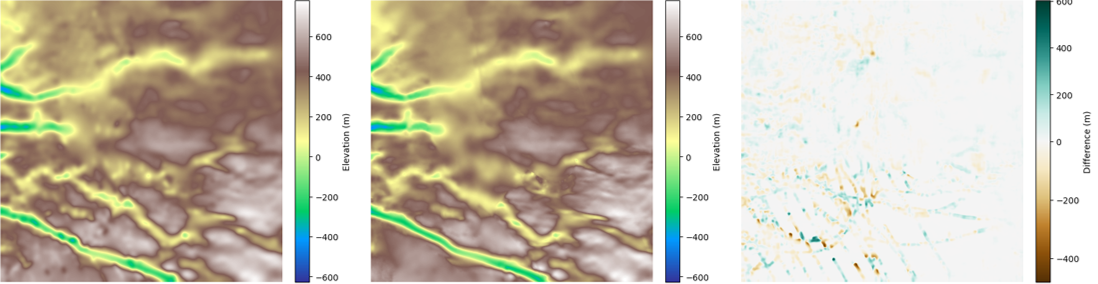}\end{minipage}\\
FPN~\cite{lin2017feature}\\
\begin{minipage}[c]{0.36\textwidth}\includegraphics[width=1\textwidth]{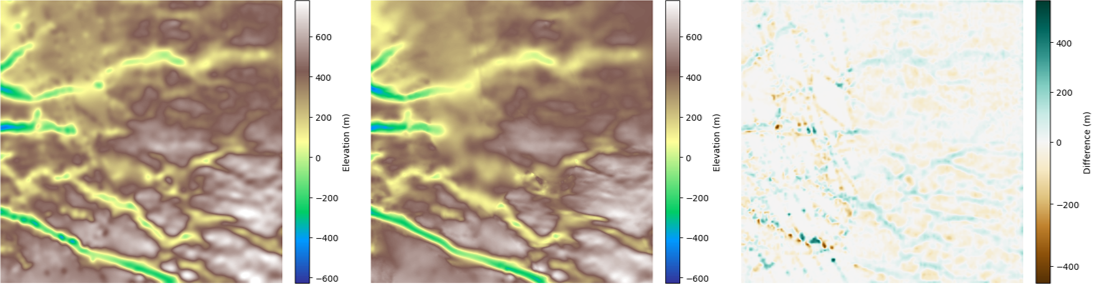}\end{minipage}&\begin{minipage}[c]{0.36\textwidth}\includegraphics[width=1\textwidth]{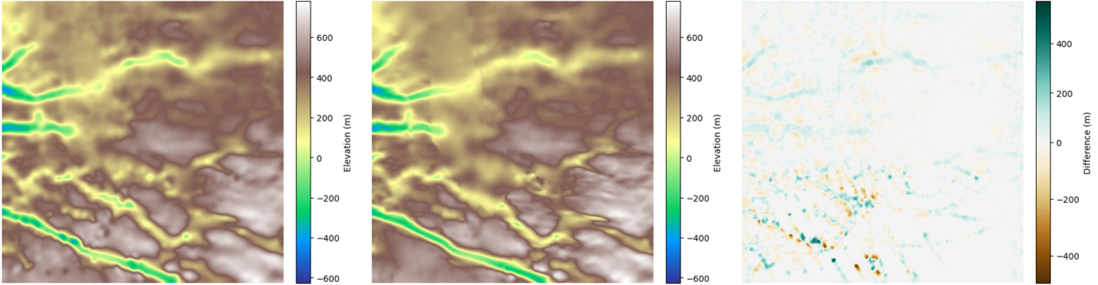}\end{minipage}\\
\textbf{Ours}\\
\begin{minipage}[c]{0.36\textwidth}\includegraphics[width=1\textwidth]{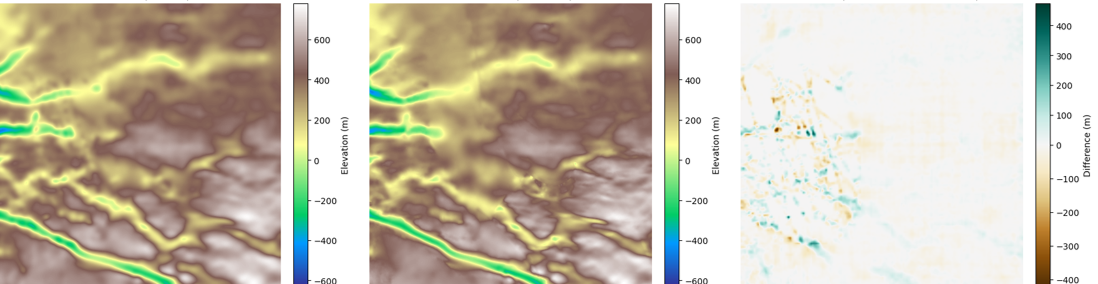}\end{minipage}&\begin{minipage}[c]{0.36\textwidth}\includegraphics[width=1\textwidth]{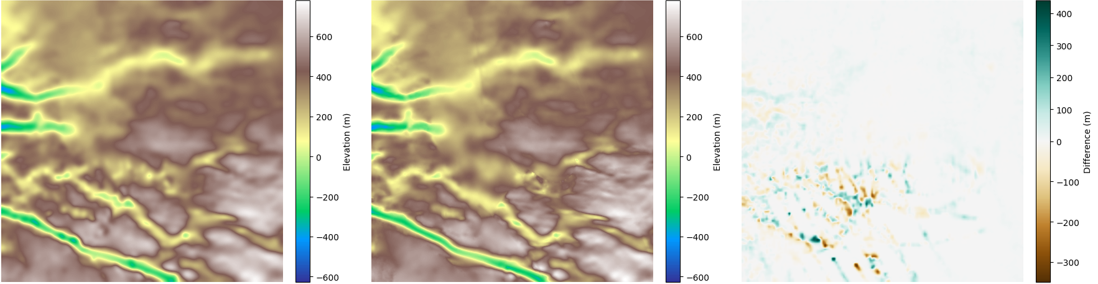}\end{minipage}\\
&\\\hline
\end{tabular}    
\end{table*}

\section{Results and Discussion}
\label{sec:results}
\subsection{Main Quantitative Comparison}
Table~\ref{tab:main_dl} reports test-core performance for two Greenland sub-regions under orthogonal block-wise hold-outs. Across two sub-regions and both block-wise splits, our physics-guided residual model is consistently best against the BedMachine prior $b_{p}$, with test-core RMSE \(3.05\)–\(10.54\) m and \(R^2=0.993\)–\(0.999\). Relative to strong CNN baselines (U-Net/Attn-U-Net/FPN/plain CNN), errors drop by roughly a factor of two on average, indicating that residual-over-prior learning with lightweight physics improves both absolute accuracy and explained variance. At held-out radar picks, simple interpolators occasionally achieve the lowest RMSE (reflecting their pass-through behavior), while our model remains competitive at picks and delivers substantially better agreement over the full test core.

Structural metrics corroborate these trends: SSIM is \(0.998\)–\(0.999\) in every split, PSNR reaches \(52.9\) dB, and roughness differences \(|\Delta\mathrm{TRI}|\) are the lowest (down to \(0.66\) in Sub-Region II and \(1.44\)–\(2.53\) in Sub-Region I). Together, the results show that the proposed design not only lowers pixel-wise error but also preserves valley/trough morphology and edge contrast while avoiding the over-smoothing seen in geostatistical baselines and the texture loss of plain CNN decoders.

\subsection{Ablation Study}
\label{sec:ablation}
Table~\ref{tab:ablation} removes one loss at a time from the full model and reports test-core metrics on both sub-regions and both splits. 

\noindent\textbf{w/o $\mathcal{L}_{\text{mass}}$:}
RMSE increases in every split (e.g., Sub-Region~I V/H: $10.54/8.12 \rightarrow 11.44/10.69$; Sub-Region~II V/H: $3.33/3.05 \rightarrow 4.68/16.10$), with the largest hit on Sub-Region~II-H, indicating mass continuity is critical where transport is highly anisotropic. \noindent\textbf{w/o $\mathcal{L}_{\text{prior}}$:} Global accuracy collapses (e.g., Sub-Region~I V/H: $10.54/8.12 \rightarrow 31.25/38.14$; Sub-Region~II V/H: $3.33/3.05 \rightarrow 21.94/24.86$) and radar-only errors also rise, confirming the need for a prior pull in data-sparse interiors. \noindent\textbf{w/o non-negativity:} Removing the $\hat{h}\!\ge\!0$ hinge yields unphysical solutions and catastrophic errors (e.g., Sub-Region~I–V RMSE $442.22$, $R^2\!<\!0$), underscoring this constraint’s necessity. \noindent\textbf{w/o $\mathcal{L}_{\text{radar}}$:} Even with physics and prior, dropping the direct fit to picks degrades agreement near observations (Sub-Region~I V/H: $10.54/8.12 \rightarrow 15.29/9.06$); Sub-Region~II is pending but expected to follow the same trend. \noindent\textbf{Takeaway:} The best performance (Table~\ref{tab:main_dl}) arises from the combination: $\mathcal{L}_{\text{radar}}$ anchors local corrections, $\mathcal{L}_{\text{prior}}$ prevents far-field drift, $\mathcal{L}_{\text{mass}}$ enforces transport-consistent structure, and non-negativity guarantees physical plausibility.

\subsection{Qualitative Maps}
\label{sec:qualitative}
Table~\ref{tab:map} visualizes prediction quality on the held-out test cores for both block-wise splits. Each row shows a method and, for each split, a triplet of panels: (i) our predicted bed, (ii) the BedMachine prior $b_p$ resampled to our grid, and (iii) the difference map (BedMachine $-\ \hat{b}$) with a zero-centered diverging colormap. The classical interpolators (IDW, Kriging) exhibit streaking along flight lines and over-smoothing of troughs. Generic CNN/FPN reduce banding but still blur fjord walls and leave structured residuals. U-Net variants recover sharper features yet retain localized artifacts (e.g., halos) and texture loss. In contrast, our physics-guided residual model preserves valley continuity and small-scale roughness while yielding low-magnitude, spatially unstructured differences, consistent across vertical and horizontal hold-outs. Note that all panels share fixed color limits; the lighter, near-zero residuals in the last row indicate closer agreement over the full test core rather than only along radar tracks.

\section{Discussion}
\label{sec:discussion}
Error patterns near fast flow and steep valleys; effect of physics losses on plausibility; radar-sparse interiors; compute/memory footprint and inference time per full grid.
Mass-conservation, flow-aware smoothing, and prior constraints act as complementary priors: the mass term reduces large-scale bias and de-striping away from radar tracks—most visibly along fast-flow corridors—while flow-aligned TV damps cross-flow wiggles yet preserves along-flow continuity; the prior-consistency term arrests far-field drift in radar-sparse interiors, and the non-negativity hinge prevents unphysical thickness. These benefits come with trade-offs: overly strong regularization can soften sharp fjord walls; regional biases in $v_x,v_y$, $\mathrm{SMB}$, or $\mathrm{\partial h/\partial t}$ can be imprinted through the mass residual; and an aggressive prior ramp may locally lock the solution to BedMachine artifacts. 

We mitigate these effects with staged weight schedules, and observe that residual errors concentrate near steep topography, tidewater margins, and zones of rapidly varying flow where inputs are uncertain. In extremely data-poor interiors, too much prior pull oversmooths roughness, whereas too little induces low-frequency drift; our leakage-safe splits show these patterns persist across vertical/horizontal domain shifts. Sensitivity analyses indicate performance depends most on prior quality and the \emph{direction} of $\mathbf{v}$ (more than magnitude), followed by $\mathrm{\partial h/\partial t}$ and SMB; Huber deltas and the confidence map produce second-order effects. Operationally, seam-free tiling bounds memory by tile size and yields cost that scales linearly with pixels (times the $8\times$ TTA factor), and EMA adds negligible overhead while improving stability for full-grid inference.

\section{Conclusion}
\label{sec:conclusion}
We presented a residual-over-prior framework for subglacial bed mapping that couples a DeepLabV3$+$ predictor with lightweight physics (mass conservation, flow-aware smoothness) and evaluates under leakage-safe spatial hold-outs. Across two Greenland sub-regions, the approach achieves strong agreement with BedMachine and preserves valley/roughness structure while degrading gracefully with distance from radar. Future work will expand to additional regions (including Antarctica), incorporate uncertainty (ensembles or Bayesian heads), explore alternative/learned priors, and tighten physics with momentum-balance surrogates or hybrid invert-and-learn schemes, while keeping evaluation leakage-safe.

\section*{Acknowledgment}
This research has been funded by the NSF HDR Institute for Harnessing Data and Model Revolution in the Polar Regions (iHARP), Award \#2118285.

{\small
\bibliographystyle{ieeenat_fullname}
\bibliography{main}
}

\clearpage
\appendix
\onecolumn
\section{Supplementary}
\subsection{Implementation Details}
\label{sup:impldet}

\paragraph{Preprocessing and features.}
All rasters are reprojected to EPSG:3413 and resampled to $\Delta\!\approx\!150$\,m on an $H{\times}W$ grid (two $600{\times}600$ extracts). Scalars are standardized per–channel using training-region statistics (mean/STD over valid land pixels). We append: (i) $\nabla s = (\partial_x s,\partial_y s)$ via central differences; (ii) Fourier coordinate features with $L{=}3$ bands on $(x,y)$; (iii) prior thickness $h_p{=}s{-}b_p$. Residual normalization uses robust statistics from training radar residuals: $\mu_t{=}\operatorname{median}$, $\sigma_t{=}1.4826\cdot\operatorname{MAD}$.

\paragraph{Radar splat and confidence.}
Radar picks are splatted to the grid using a cKDTree with $K{=}9$ nearest cells and Gaussian weights $\exp\!\big(-\!(d/r)^2\big)$ where $r{=}\,2.5$\,px${}\times\Delta$. The radar confidence map is $c(x,y){=}\exp(-d/\tau)$ with $\tau{=}12$\,px (grid distance to nearest pick). Weights: $w_{\text{radar}}{=}\max(\varepsilon,c)$, physics losses use $(1{-}c)$, prior loss uses $(1{-}c)^2$ and is masked at picks. To avoid pulling on steep slopes, we attenuate $L_{\text{prior}}$ by a slope weight $w_{\nabla b_p}{=}\exp(-\|\nabla b_p\|/s_{90})$ where $s_{90}$ is the 90th percentile of $\|\nabla b_p\|$.

\paragraph{Model and heads.}
DeepLabV3$+$ decoder with a ResNet-50 encoder (output stride 16). Low-level projection $1{\times}1$ (48 ch), ASPP rates $(1,6,12,18)$, GroupNorm everywhere, LeakyReLU, dropout $0.1$ in ASPP/decoder. Output is the normalized residual $\,\hat r\in\mathbb{R}^{H\times W}$.

\paragraph{Losses and weights.}
Total loss $L=\lambda_{\text{data}}L_{\text{radar}}+\lambda_{\text{phys}}L_{\text{mass}}+\lambda_{\text{tv}}L_{\text{flowTV}}+\lambda_{\text{lap}}L_{\text{lap}}+\lambda_{\ge0}L_{\ge0}+\lambda_{\text{prior}}L_{\text{prior}}$.
Huber deltas: $\delta_{\text{radar}}{=}1.0$, $\delta_{\text{mass}}{=}5.0$, $\delta_{\text{prior}}{=}10.0$.
Flow-aligned TV uses unit flow $\mathbf{u}=\mathbf{v}/(\|\mathbf{v}\|{+}\epsilon)$ with
$L_{\text{flowTV}}=\beta_{\perp}\|\nabla \hat h\!\cdot\!\mathbf{u}_{\perp}\|_1+\beta_{\parallel}\|\nabla \hat h\!\cdot\!\mathbf{u}\|_1$,
$\beta_{\perp}{=}0.9$, $\beta_{\parallel}{=}0.35$.
Laplacian uses the $3{\times}3$ kernel $\begin{psmallmatrix}0&-1&0\\-1&4&-1\\0&-1&0\end{psmallmatrix}$ on $\hat r$.
Mass-conservation residual is applied multi-scale at pooling factors $\{1,2,4\}$, with Gaussian-smoothed fluxes; kernel schedule: first half of training uses size $11$, $\sigma{=}3.5$, then size $15$, $\sigma{=}5.0$.

\paragraph{Schedules (ramp and weights).}
Unless noted: $\lambda_{\text{data}}{=}2.0$, $\lambda_{\text{phys}}{=}10^{-2}$ (linear ramp $0{\rightarrow}$target over the first $\sim\!90\%$ of epochs), $\lambda_{\text{tv}}{=}5{\times}10^{-4}$, $\lambda_{\text{lap}}{=}2{\times}10^{-4}$, $\lambda_{\ge0}{=}10^{-3}$, $\lambda_{\text{prior}}{=}5{\times}10^{-3}$ (ramp from 30\% to 90\% of training).

\paragraph{Optimization and training protocol.}
AdamW (lr $1{\times}10^{-4}$, weight decay $1{\times}10^{-4}$) with cosine warm restarts ($T_0{=}500$, $T_{\text{mult}}{=}2$); batch size $8$; up to $6000$ epochs with early stopping (patience $2000$ epochs) on masked radar-thickness fit inside the train core. Seed fixed to $42$. WeightedRandomSampler favors tiles that contain any radar in the patch core (weights $6{:}1$).

\paragraph{Geo-aware augmentation and inference.}
During training we apply random $\pi/2$ rotations and flips with probability $0.75$ (vector-aware transforms for $v_x,v_y$ and gradient channels). At test time we use 8-way TTA (4 rotations $\times$ horizontal flip), inverse-transform and average. EMA decay $0.999$; seam-free tiled inference with patch $256$, stride $64$, and core border $b{=}96$\,px. For leakage-safe splits we erode train/test blocks by $\delta{=}96$\,px and compute metrics strictly on the held-out test core.

\paragraph{Evaluation specifics.}
We select a single global rotation/flip that minimizes RMSE against the reference once per split, then compute MAE/RMSE/$R^2$, SSIM, and PSNR (range $=$ dynamic range of the reference in the core). TRI uses a $3{\times}3$ neighborhood; $|\Delta\mathrm{TRI}|$ is mean absolute difference in the core. Radar-only errors sample $\hat b$ at held-out picks in the core.

\subsection{Distance-to-Radar Stratification}
\label{sec:disradar}
\begin{figure*}[ht!]
\centering
\begin{tikzpicture}
\begin{groupplot}[
  group style={
    group size=2 by 2,
    horizontal sep=16mm,
    vertical sep=24mm
  },
  width=0.42\textwidth,
  height=0.33\textwidth,
  ymin=0, ymajorgrids,
  symbolic x coords={0-2,2-6,6-inf},
  xtick=data,
  xticklabels={$0\!-\!2$,$2\!-\!6$,$6\!-\!\infty$},
  every axis plot/.append style={mark size=3.3pt, line width=1.2pt},
  xlabel={Distance to radar (px)},
  ylabel={RMSE (m)}
]

\nextgroupplot[
  title={Sub-Region I — Vertical},
  ymax=50,
  legend to name=SharedLegend,
  legend columns=5,
  legend style={draw=none, /tikz/every even column/.append style={column sep=0.6em}, /tikz/every odd column/.append style={column sep=0.6em}}
]
\addplot+[very thick, mark=triangle, color=blue]         coordinates {(0-2,30.20) (2-6,29.64) (6-inf,29.25)}; \addlegendentry{CNN}
\addplot+[very thick, mark=square,  color=green]         coordinates {(0-2,34.64) (2-6,27.63) (6-inf,20.61)}; \addlegendentry{U-Net}
\addplot+[very thick, mark=diamond, color=orange]        coordinates {(0-2,30.57) (2-6,25.81) (6-inf,21.64)}; \addlegendentry{Att.\ U\text{-}Net}
\addplot+[very thick, mark=x,        color=purple]        coordinates {(0-2,44.07) (2-6,38.61) (6-inf,32.84)}; \addlegendentry{FPN}
\addplot+[very thick, mark=o,        color=black]         coordinates {(0-2,10.94) (2-6,11.17) (6-inf,10.21)}; \addlegendentry{Ours}

\nextgroupplot[title={Sub-Region I — Horizontal}, ymax=50, ylabel={}]
\addplot+[very thick, mark=triangle, color=blue,          forget plot] coordinates {(0-2,24.71) (2-6,24.13) (6-inf,22.47)};
\addplot+[very thick, mark=square,  color=green,           forget plot] coordinates {(0-2,19.83) (2-6,20.85) (6-inf,16.87)};
\addplot+[very thick, mark=diamond, color=orange,         forget plot] coordinates {(0-2,22.03) (2-6,21.07) (6-inf,20.31)};
\addplot+[very thick, mark=x,        color=purple,         forget plot] coordinates {(0-2,32.12) (2-6,29.82) (6-inf,24.29)};
\addplot+[very thick, mark=o,        color=black,          forget plot] coordinates {(0-2,9.33) (2-6,8.91) (6-inf,7.48)};

\nextgroupplot[title={Sub-Region II — Vertical}, ymax=16]
\addplot+[very thick, mark=triangle, color=blue,          forget plot] coordinates {(0-2,14.26) (2-6,14.08) (6-inf,9.51)};
\addplot+[very thick, mark=square,  color=green,           forget plot] coordinates {(0-2,9.89) (2-6,9.57) (6-inf,6.14)};
\addplot+[very thick, mark=diamond, color=orange,         forget plot] coordinates {(0-2,10.77) (2-6,9.91) (6-inf,7.57)};
\addplot+[very thick, mark=x,        color=purple,         forget plot] coordinates {(0-2,14.52) (2-6,14.84) (6-inf,10.37)};
\addplot+[very thick, mark=o,        color=black,          forget plot] coordinates {(0-2,5.75) (2-6,4.84) (6-inf,2.76)};

\nextgroupplot[title={Sub-Region II — Horizontal}, ymax=16, ylabel={}]
\addplot+[very thick, mark=triangle, color=blue,          forget plot] coordinates {(0-2,10.81) (2-6,11.47) (6-inf,8.69)};
\addplot+[very thick, mark=square,  color=green,          forget plot] coordinates {(0-2,14.18) (2-6,14.23) (6-inf,9.39)};
\addplot+[very thick, mark=diamond, color=orange,         forget plot] coordinates {(0-2,15.31) (2-6,15.75) (6-inf,10.86)};
\addplot+[very thick, mark=x,        color=purple,         forget plot] coordinates {(0-2,11.33) (2-6,11.27) (6-inf,9.74)};
\addplot+[very thick, mark=o,        color=black,          forget plot] coordinates {(0-2,3.44) (2-6,3.36) (6-inf,2.91)};

\end{groupplot}
\end{tikzpicture}
\smallskip
\pgfplotslegendfromname{SharedLegend}

\caption{\textbf{RMSE vs.\ distance to radar on held-out test cores.}
For each sub-region (rows) and split (columns), RMSE is reported in three distance bins (px) from the nearest radar pick: $0\!-\!2$, $2\!-\!6$, and $6\!-\!\infty$. Lines encode methods (markers/colors in the legend). Lower is better.}
\label{fig:rmse_vs_distance}
\end{figure*}
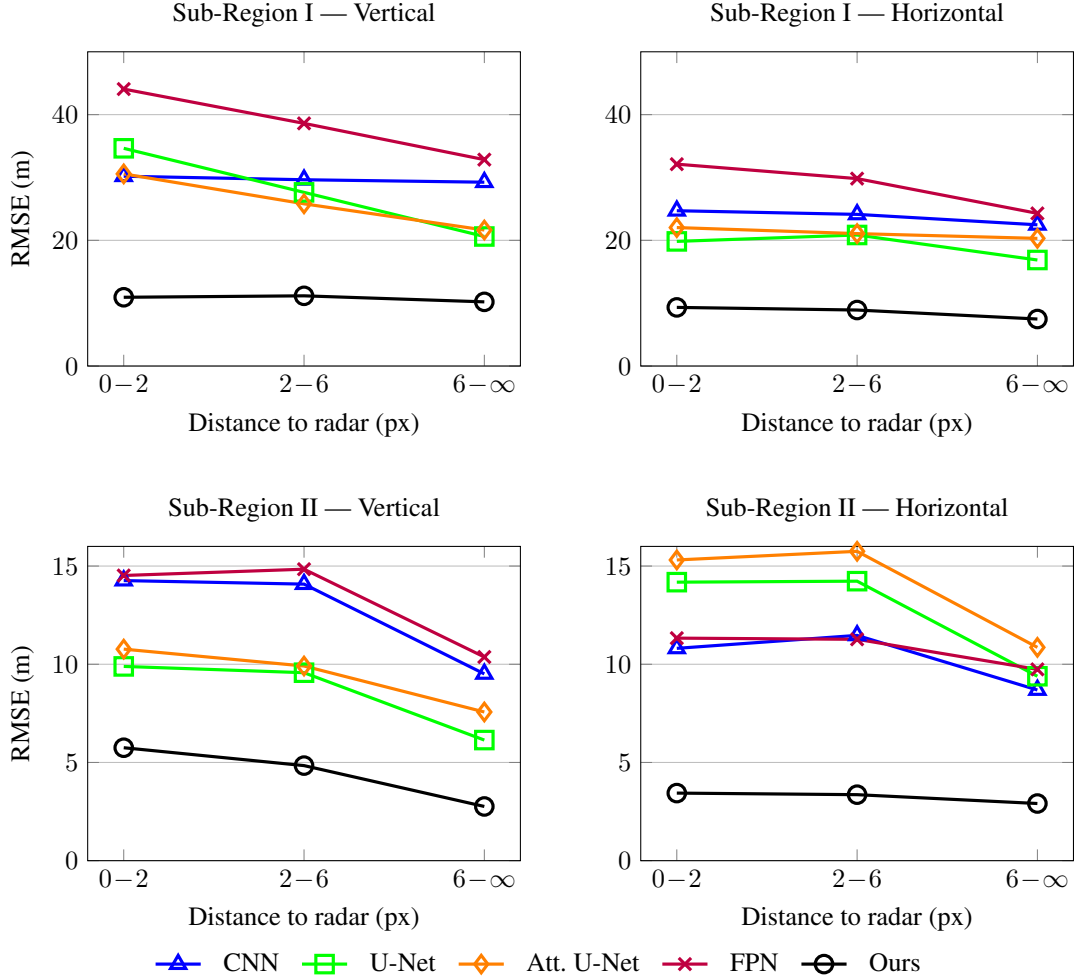

To examine how performance varies with observational support, we stratify test-core pixels by distance to the nearest radar pick into three bins: $0$–$2$, $2$–$6$, and $>6$ pixels (1 px $\approx 150$ m), and report RMSE in each bin (Fig.~\ref{fig:rmse_vs_distance}). Across both sub-regions and both splits, our physics-guided residual model attains the lowest RMSE in \emph{every} bin and degrades the least as distance increases. In fact, RMSE typically \emph{decreases} farther from picks for our method (e.g., Sub-Region~I—H: $9.33\!\rightarrow\!8.91\!\rightarrow\!7.48$ m; Sub-Region~II—V: $5.75\!\rightarrow\!4.84\!\rightarrow\!2.76$ m), consistent with the design: near margins and complex flow (where picks cluster) the field is harder to predict, whereas in radar-sparse interiors the prior-consistency and mass terms stabilize the reconstruction. By contrast, CNN/U-Net/FPN exhibit larger near-pick errors and a shallower improvement with distance (e.g., Sub-Region~I—H at $>6$ px: U-Net/FPN $16.87/24.29$ m vs.\ ours $7.48$ m; Sub-Region~II—V at $>6$ px: U-Net/FPN $6.14/10.37$ m vs.\ ours $2.76$ m). These trends indicate that residual-over-prior learning with lightweight physics yields robust generalization in radar-sparse interiors while avoiding the banding and over-smoothing seen in non-physics baselines. (Note: stratification uses BedMachine as the reference).

\subsection{Ablation Study Using Feature Pyramid Network (FPN) as a Backbone}
\label{sec:supp:ablate-fpn}
Table~\ref{tab:ablation_fpn} repeats the loss-component ablation with an FPN decoder in place of DeepLabV3$+$ while keeping the same residual-over-prior target, inputs, schedules, and leakage-safe protocol. The trends mirror the main-paper ablation: (i) \textbf{Prior-consistency} is pivotal—removing $\mathcal{L}_{\text{prior}}$ produces the largest degradation on the BedMachine comparison across splits (e.g., Sub-Region~I V/H: RMSE $46.09/39.77$\,m; Sub-Region~II V: $70.05$\,m with $R^{2}{<}0$), confirming that an explicit pull toward the prior is required in radar-sparse interiors. (ii) The \textbf{non-negativity} hinge is essential for physical plausibility; without it, errors are catastrophic with strongly negative $R^{2}$ in all splits (e.g., Sub-Region~I V: RMSE $381.92$\,m). (iii) The \textbf{mass-conservation} term improves field structure—particularly in anisotropic splits—though its removal yields moderate increases relative to (i)/(ii) (e.g., Sub-Region~I V/H: $35.90/21.50$\,m RMSE). (iv) Dropping the \textbf{radar data term} $\mathcal{L}_{\text{radar}}$ raises pick-proximal errors (e.g., Sub-Region~I V: RMSE $84.97$\,m vs.\ with-fit) and modestly worsens BedMachine agreement, indicating that $\mathcal{L}_{\text{radar}}$ anchors local corrections while the physics and prior govern global behavior.

Overall, the \emph{ranking} of term importance (non-negativity/prior $\gg$ mass $\gtrsim$ radar) is consistent with the DeepLabV3$+$ results in the main paper, while absolute errors are somewhat larger with FPN—reflecting decoder capacity rather than a change in the efficacy of the loss design. These findings suggest the proposed residual+physics formulation is \emph{backbone-agnostic}: the same components deliver the same qualitative benefits across architectures. 

\begin{table*}[ht!]
\centering
\caption{\textbf{Ablations of loss components on held-out test cores using FPN as a backbone.}
Effect of removing mass-conservation (\(\mathcal{L}_{\text{mass}}\)), prior-consistency (\(\mathcal{L}_{\text{prior}}\)), non-negativity, and radar data fit (\(\mathcal{L}_{\text{radar}}\)). Protocol matches the main-paper ablation (residual-over-prior target, identical schedules, and test-core scoring); values are in meters.}

\label{tab:ablation_fpn}
\begin{adjustbox}{width=\textwidth}
\begin{tabular}{l|c|cccccc|cccccc}
\hline
\multirow{3}{*}{Method} & \multirow{3}{*}{Reference Data} & \multicolumn{6}{c|}{Sub-Region I}                              & \multicolumn{6}{c}{Sub-Region II}                             \\\cline{3-14}
                       &                                 & \multicolumn{3}{c}{Vertical} & \multicolumn{3}{c|}{Horizontal} & \multicolumn{3}{c}{Vertical} & \multicolumn{3}{c}{Horizontal} \\\cline{3-14}
&& \multicolumn{1}{c}{MAE$\downarrow$} & \multicolumn{1}{c}{RMSE$\downarrow$} & \multicolumn{1}{c}{{R$^{2}\uparrow$}} & \multicolumn{1}{c}{MAE$\downarrow$} & \multicolumn{1}{c}{RMSE$\downarrow$} & \multicolumn{1}{c|}{R$^{2}\uparrow$} & \multicolumn{1}{c}{MAE$\downarrow$} & \multicolumn{1}{c}{RMSE$\downarrow$} & \multicolumn{1}{c}{R$^{2}\uparrow$} & \multicolumn{1}{c}{MAE$\downarrow$} & \multicolumn{1}{c}{RMSE$\downarrow$} & \multicolumn{1}{c}{{R$^{2}\uparrow$}} \\
\hline
\multirow{2}{*}{w/o $\mathcal{L}_{mass}$}&BedMachine&25.98&35.90&0.920&14.50&21.50&0.970&6.20&9.53&0.954&6.33&10.44&0.985\\
& Radar&76.77&96.41&0.421&93.32&123.89&0.265&99.43&128.00&-0.020&59.11&89.63&0.447\\\hline
\multirow{2}{*}{w/o $\mathcal{L}_{prior}$}&BedMachine&35.88&46.09&0.868&30.17&39.77&0.896&54.62&70.05&-1.483&25.46&30.52&0.868\\
& Radar&76.63&98.38&0.397&89.92&117.36&0.340&109.11&140.86&-0.235&61.98&94.40&0.387\\\hline
\multirow{2}{*}{w/o non-negativity}&BedMachine&375.73&381.92&-8.093&108.13&124.96&-0.024&280.11&287.00&-40.688&87.28&97.55&-0.344\\
& Radar&406.45&422.31&-10.105&154.17&191.96&-0.766&330.30&361.64&-7.143&112.66&147.55&-0.499\\\hline
\multirow{2}{*}{w/o $\mathcal{L}_{radar}$}&BedMachine&22.15&31.17&0.939&14.54&21.36&0.970&6.84&9.31&0.956&5.15&8.32&0.990\\
& Radar&65.28&84.97&0.550&92.32&121.24&0.296&97.76&126.56&0.003&58.73&88.72&0.458\\\hline
\end{tabular}
\end{adjustbox}
\end{table*}

\subsection{Quality Maps for Sub-Region II}
All panels use the same elevation limits; difference maps use a zero-centered diverging colormap (white $\approx 0$\,m). A single rotation/flip alignment is fixed once per split (as in the main protocol).

\begin{table*}[ht!]
\caption{\textbf{Qualitative comparison on held-out test cores for Sub-region II.}
For each method (rows) and split (columns: \emph{Vertical}, \emph{Horizontal}), we show triplets:
\emph{Prediction} $\mid$ \emph{Prior} $b_p$ (BedMachine) $\mid$ \emph{Difference} (BedMachine $-\ \hat{b}$). All panels use consistent color limits; difference maps use a zero-centered diverging colormap (white $\approx$ 0 m).}
\label{tab:map_subreg_II}
\centering
\begin{tabular}{l|l}
\hline
Vertical&Horizontal\\ \hline
\multicolumn{2}{c}{Prediction $\mid$ Prior $b_{p}$ $\mid$ Difference}\\\hline
IDW~\cite{shepard1968two}\\
\begin{minipage}[c]{0.48\textwidth}\includegraphics[width=1\textwidth]{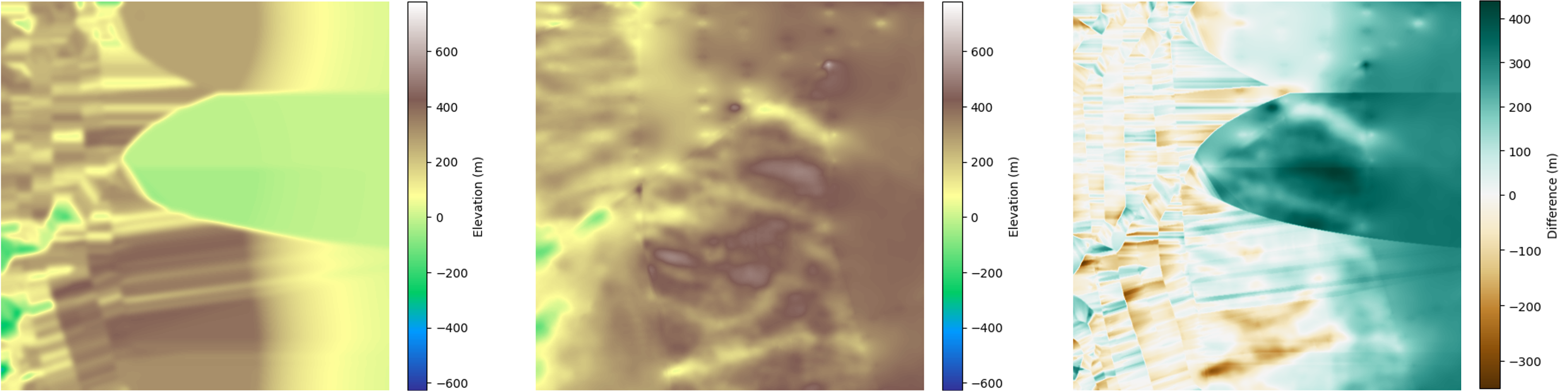}\end{minipage}&\begin{minipage}[c]{0.48\textwidth}\includegraphics[width=1\textwidth]{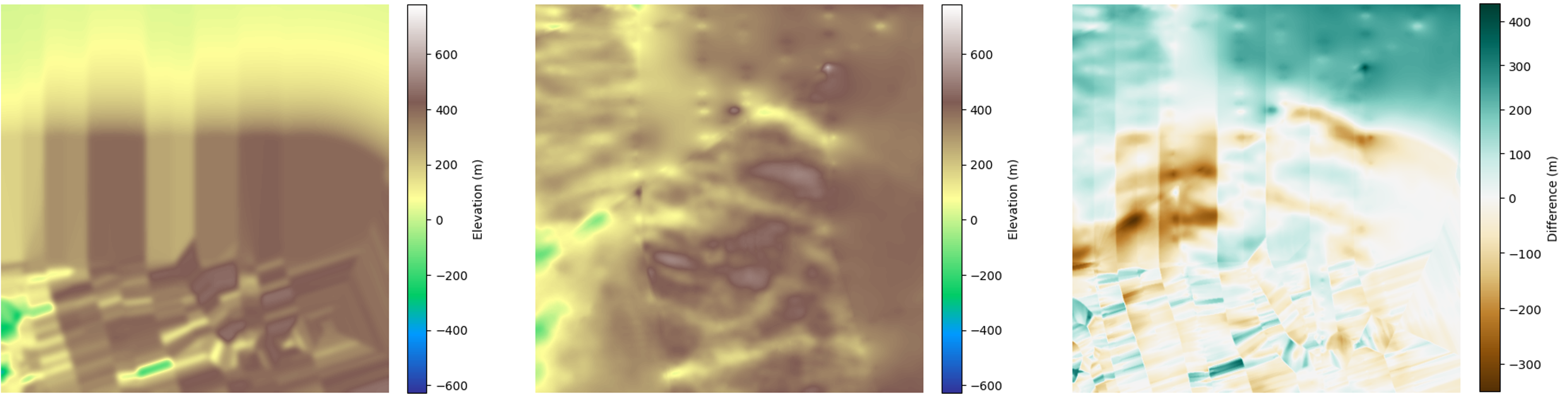}\end{minipage}\\
Kriging~\cite{cressie1985fitting,Hengl2007RegressionKriging,cressie1980robust}\\
\begin{minipage}[c]{0.48\textwidth}\includegraphics[width=1\textwidth]{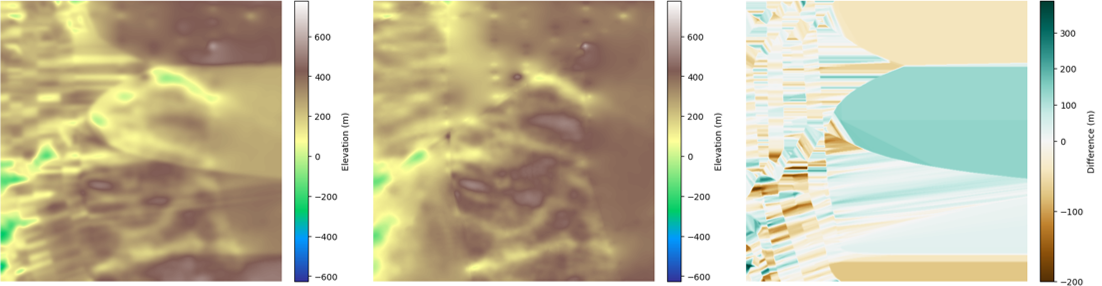}\end{minipage}&\begin{minipage}[c]{0.48\textwidth}\includegraphics[width=1\textwidth]{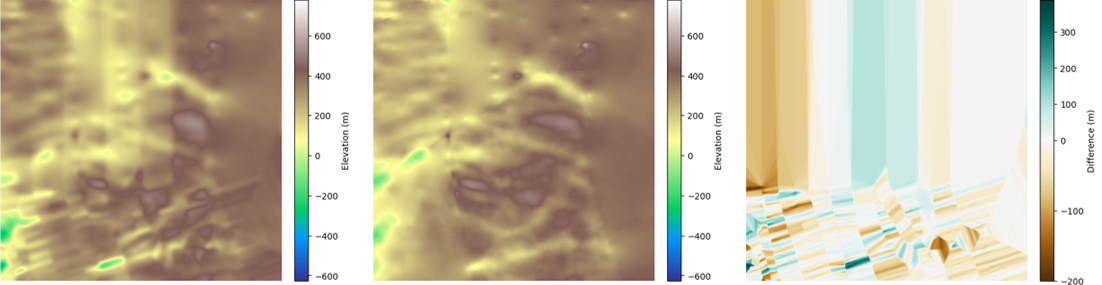}\end{minipage}\\
CNN~\cite{he2016deep,wu2018group,yu2017dilated}\\
\begin{minipage}[c]{0.48\textwidth}\includegraphics[width=1\textwidth]{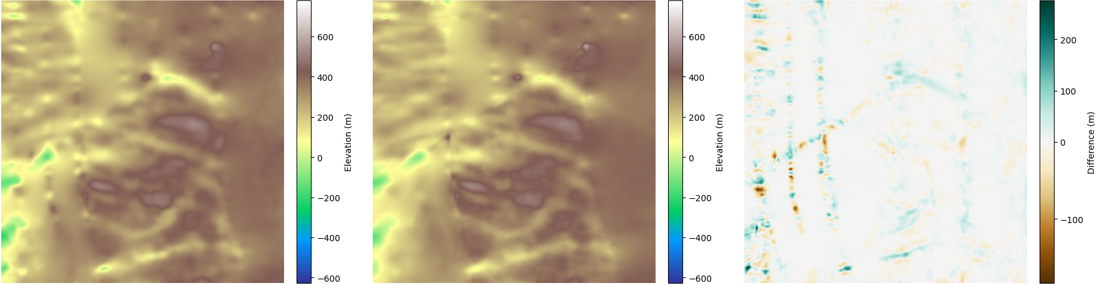}\end{minipage}&\begin{minipage}[c]{0.48\textwidth}\includegraphics[width=1\textwidth]{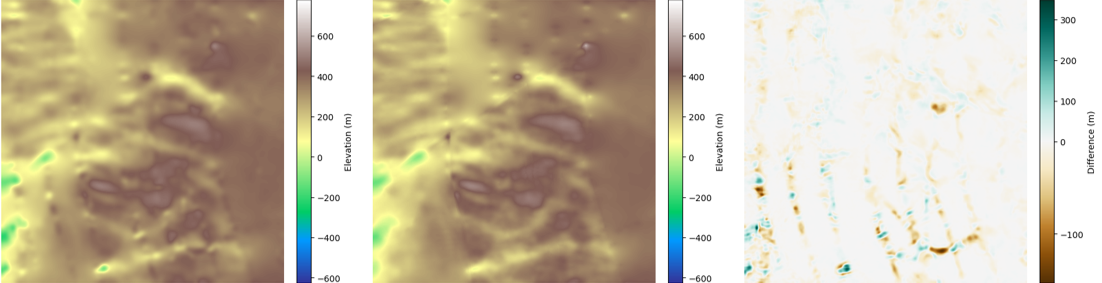}\end{minipage}\\
U-Net~\cite{ronneberger2015u}\\
\begin{minipage}[c]{0.48\textwidth}\includegraphics[width=1\textwidth]{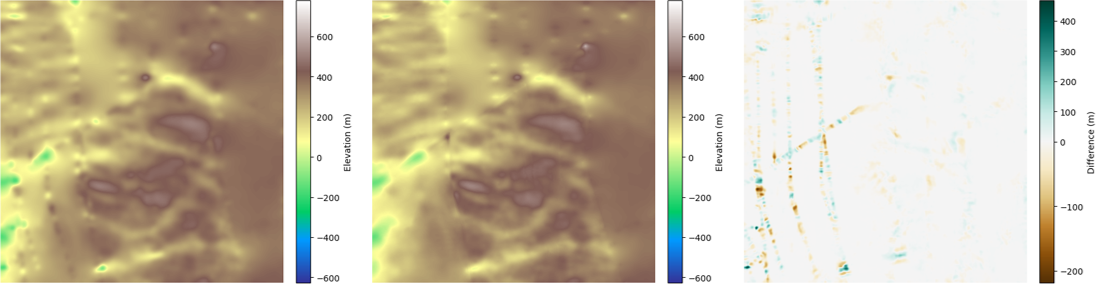}\end{minipage}&\begin{minipage}[c]{0.48\textwidth}\includegraphics[width=1\textwidth]{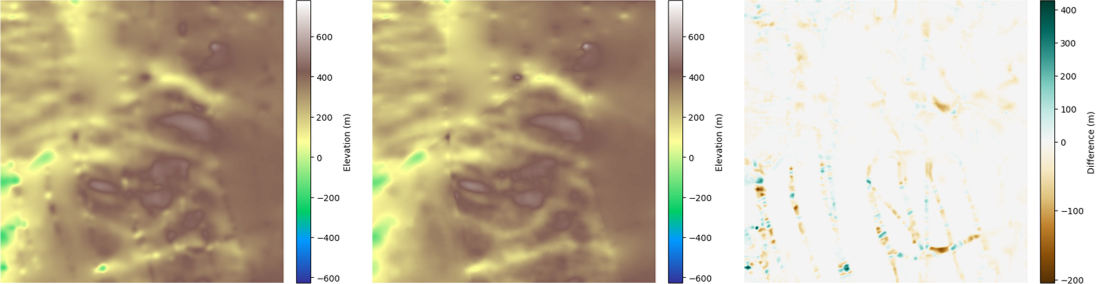}\end{minipage}\\
&\\\hline
\end{tabular}    
\end{table*}
\begin{center}
\begin{tabular}{l|l}
\hline
Vertical&Horizontal\\ \hline
\multicolumn{2}{c}{Prediction $\mid$ Prior $b_{p}$ $\mid$ Difference}\\\hline
Att. U-Net~\cite{ronneberger2015u,oktay2018attention}\\
\begin{minipage}[c]{0.48\textwidth}\includegraphics[width=1\textwidth]{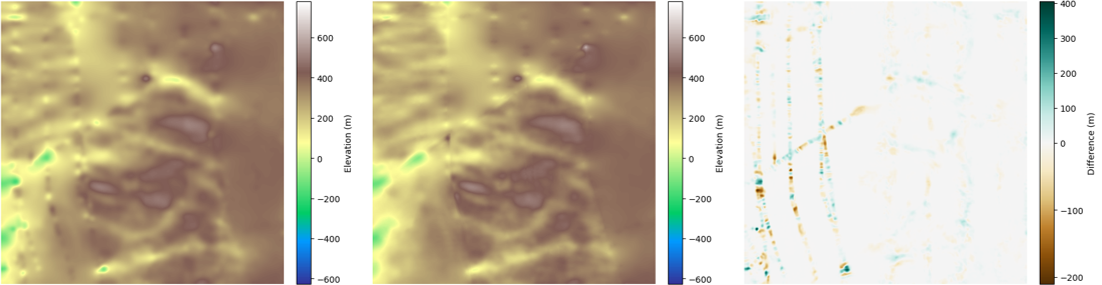}\end{minipage}&\begin{minipage}[c]{0.48\textwidth}\includegraphics[width=1\textwidth]{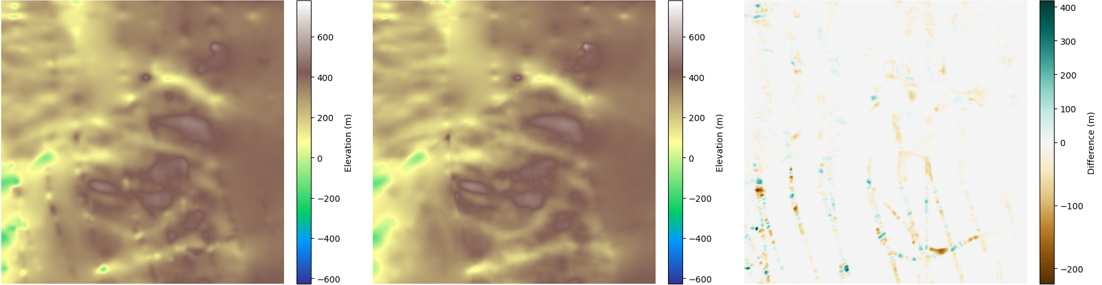}\end{minipage}\\
FPN~\cite{lin2017feature}\\
\begin{minipage}[c]{0.48\textwidth}\includegraphics[width=1\textwidth]{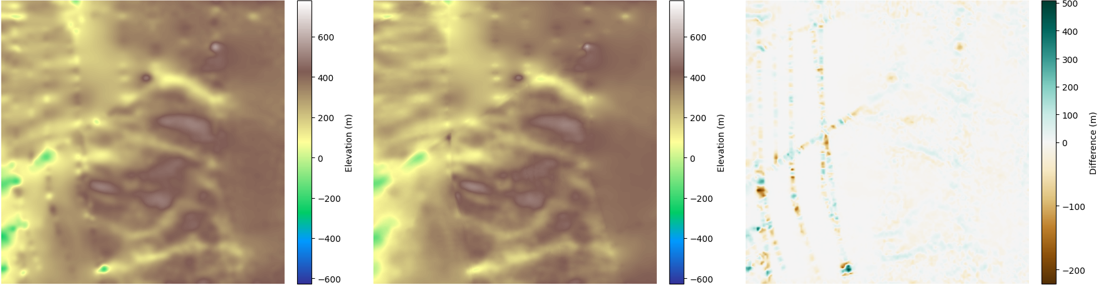}\end{minipage}&\begin{minipage}[c]{0.48\textwidth}\includegraphics[width=1\textwidth]{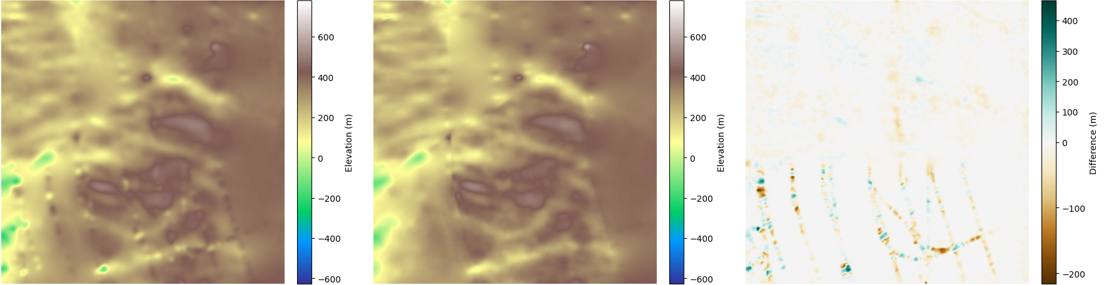}\end{minipage}\\
\textbf{Ours}\\
\begin{minipage}[c]{0.48\textwidth}\includegraphics[width=1\textwidth]{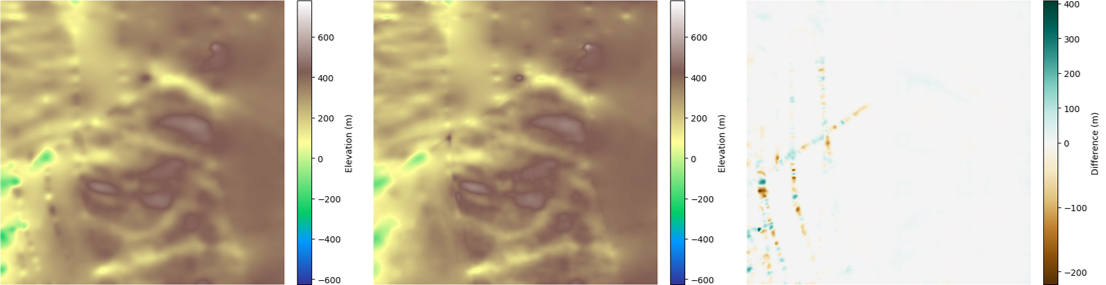}\end{minipage}&\begin{minipage}[c]{0.48\textwidth}\includegraphics[width=1\textwidth]{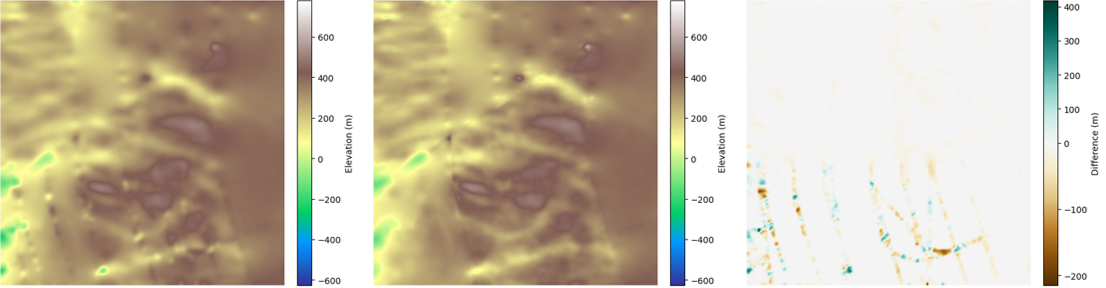}\end{minipage}\\
&\\\hline
\end{tabular}    
\end{center}

Classical interpolators (IDW, Kriging) either over-smooth fjord walls and interior ridges or exhibit track-aligned streaks, resulting in structured residuals in the difference maps. CNN/U-Net/FPN reduce banding but still blur valley edges and leave halo-like artifacts near steep margins. In contrast, our physics-guided residual model preserves trough continuity and flank sharpness while producing low-amplitude, spatially diffuse differences across both vertical and horizontal splits. These visuals align with the quantitative trends reported for Sub-region~II: very high structural fidelity (SSIM $\approx 0.999$, PSNR up to $\sim\!52.94$\,dB) and the lowest roughness discrepancy ($|\Delta\mathrm{TRI}| \approx 0.66$), together with low test-core RMSE (3.33\,m vertical, 3.05\,m horizontal). Notably, residuals for our method remain decorrelated from radar track geometry, indicating that corrections are driven by the residual-over-prior + physics design rather than memorization of pick patterns.

\subsection{Comparative Assessment with Classical/ML Baselines}
\label{sec:ml_comparison}

\begin{table*}[ht!]
\centering
\caption{\textbf{Machine-learning baselines on held-out test cores.}
Ridge, ElasticNet, KNN, Random Forest (RF), Gradient Boosting (GB), and Support Vector Regression (SVR) trained on the same inputs and residual target as our model. “BedMachine” rows score agreement with the BedMachine prior $b_{p}$; “Radar” rows report errors at held-out radar picks. Protocol matches the main paper (block-wise splits with safety buffer, single orientation alignment per split, metrics on the test core). Values are in meters; higher $R^2$ is better.}
\label{tab:ml_comparison}
\begin{adjustbox}{width=\textwidth}
\begin{tabular}{l|c|cccccc|cccccc}
\hline
\multirow{3}{*}{Method} & \multirow{3}{*}{Reference Data} & \multicolumn{6}{c|}{Sub-Region I}                              & \multicolumn{6}{c}{Sub-Region II}                             \\\cline{3-14}
                       &                                 & \multicolumn{3}{c}{Vertical} & \multicolumn{3}{c|}{Horizontal} & \multicolumn{3}{c}{Vertical} & \multicolumn{3}{c}{Horizontal} \\\cline{3-14}
&& \multicolumn{1}{c}{MAE$\downarrow$} & \multicolumn{1}{c}{RMSE$\downarrow$} & \multicolumn{1}{c}{{R$^{2}\uparrow$}} & \multicolumn{1}{c}{MAE$\downarrow$} & \multicolumn{1}{c}{RMSE$\downarrow$} & \multicolumn{1}{c|}{R$^{2}\uparrow$} & \multicolumn{1}{c}{MAE$\downarrow$} & \multicolumn{1}{c}{RMSE$\downarrow$} & \multicolumn{1}{c}{R$^{2}\uparrow$} & \multicolumn{1}{c}{MAE$\downarrow$} & \multicolumn{1}{c}{RMSE$\downarrow$} & \multicolumn{1}{c}{{R$^{2}\uparrow$}} \\
\hline
\multirow{2}{*}{Ridge}&BedMachine&870.54&1027.53&-64.819&731.80&957.01&-59.064&2813.93&3164.69&-5067.668&537.27&725.08&-73.240\\
& Radar&664.39&893.91&-48.753&252.95&467.66&-9.478&2153.17&2705.44&-454.737&24.50&704.96&-33.208\\\hline
\multirow{2}{*}{ElasticNet}&BedMachine&80.01 &88.79&0.508 &95.16&105.61&0.269&145.14&148.68&-10.188&36.67&48.13&0.673\\
& Radar&89.65&105.59&0.306&103.88&129.35&0.198&142.29&172.32&-0.849&79.75&115.22&0.086\\\hline
\multirow{2}{*}{KNN}&BedMachine&94.07&128.69&-0.032&60.36&90.23&0.466&46.97&61.71&-0.927&71.68&114.78  &-0.860\\
& Radar&101.17&134.56&-0.127&120.57&154.78&-0.148&97.78&125.95&0.012&110.02&157.71&-0.712\\\hline
\multirow{2}{*}{Random Forest}&BedMachine&86.88&100.75&0.367&64.68&86.52&0.509&47.42&65.73&-1.187&42.64&58.79&0.512\\
& Radar&96.96&117.68&0.138&98.26&124.22&0.261&114.36&144.92&-0.308&65.23&86.66&0.483\\\hline
\multirow{2}{*}{Gradient Boosting}&BedMachine&152.22&157.13&-0.539&34.49&47.68&0.851&61.93&71.18&-1.564&30.15&39.93&0.775\\
& Radar&114.84&135.03&-0.135&114.54&154.25&-0.140&107.59&135.90&-0.150&69.72&94.91&0.380\\\hline
\multirow{2}{*}{SVR}&BedMachine&28.02&28.53&0.949&25.6&33.16&0.928&14.85&15.19&0.883&16.95&25.27&0.910\\
& Radar&75.82&96.11&0.425&96.16&121.86&0.289&97.77&124.61&0.033&64.72&97.38&0.347\\\hline
\end{tabular}
\end{adjustbox}
\end{table*}

Table~\ref{tab:ml_comparison} reports leakage-safe test-core performance for common machine-learning regressors trained on the same inputs (surface, velocity, SMB, $\mathrm{dhdt}$, gradients, Fourier coords, prior thickness) and residual target as in the main paper. We follow the identical protocol: orthogonal block-wise splits, receptive-field buffers, single rotation/flip chosen once per split, and scoring on the held-out core. Hyperparameters were tuned on a slice of the train core to avoid test leakage. Overall, support-vector regression (SVR) is the strongest ML baseline, attaining RMSE $\sim$25–33\,m on Sub-region~I and $\sim$15–25\,m on Sub-region~II against BedMachine, while linear (Ridge/ElasticNet) and tree ensembles (RF/GB) lag or overfit. KNN performs inconsistently across splits. Compared to the DeepLabV3$+$ residual+physics model in the main paper (RMSE 3–11\,m), these ML baselines exhibit substantially higher errors and lower $R^2$, underscoring the value of: (i) learning \emph{residual thickness} over a prior, (ii) physics-guided regularization, and (iii) leakage-safe training/evaluation.

\end{document}